\documentclass[twocolumn]{IEEEtran}
\usepackage[T1]{fontenc}
\usepackage{cite}
\usepackage{textcomp}
\usepackage{dblfloatfix}
\usepackage{amsmath,amssymb,amsfonts}
\usepackage{algorithmic}
\usepackage{graphicx}
\usepackage{textcomp}
\usepackage{booktabs}
\usepackage{graphicx}
\usepackage{multirow}
\usepackage{subfig}
\usepackage{array}
\usepackage{mathtext}
\usepackage{enumitem}
\usepackage{amsmath}
\usepackage{algorithmic}
\usepackage{algorithm}
\usepackage{epsfig}
\usepackage{url}
\usepackage{epstopdf}
\usepackage{cite}
\usepackage{setspace}
\usepackage{xcolor}
\usepackage{hyperref}

\usepackage{caption}
\usepackage{tabularray}
\UseTblrLibrary{booktabs}

\renewcommand{\v}[1]{{\mathbf{#1}}}

\newcommand\MyBox[2]{
  \fbox{\lower0.75cm
    \vbox to 1.7cm{\vfil
      \hbox to 1.7cm{\hfil\parbox{1.4cm}{#1\\#2}\hfil}
      \vfil}%
  }%
}

\usepackage{xcolor}
\usepackage{subfig}

\def\BibTeX{{\rm B\kern-.05em{\sc i\kern-.025em b}\kern-.08em
    T\kern-.1667em\lower.7ex\hbox{E}\kern-.125emX}}
\makeatletter
\def\ps@IEEEtitlepagestyle{%
  \def\@oddfoot{\mycopyrightnotice}%
  \def\@oddhead{\hbox{}\@IEEEheaderstyle\leftmark\hfil\thepage}\relax
  \def\@evenhead{\@IEEEheaderstyle\thepage\hfil\leftmark\hbox{}}\relax
  \def\@evenfoot{}%
}
\def\mycopyrightnotice{%
  \begin{minipage}{\textwidth}
  \centering \scriptsize
  Copyright~\copyright~2022 IEEE. Personal use of this material is permitted. Permission from IEEE must be obtained for all other uses, in any current or future media, including\\reprinting/republishing this material for advertising or promotional purposes, creating new collective works, for resale or redistribution to servers or lists, or reuse of any copyrighted component of this work in other works by sending a request to pubs-permissions@ieee.org.
  \end{minipage}
}
\makeatother
\begin{document}
\pagenumbering{roman}
\bibliographystyle{ieeetr}

\title{Are You Comfortable Now: Deep Learning the Temporal Variation in Thermal Comfort in Winters}
\author{\IEEEauthorblockN{Betty Lala$^\Phi$, Srikant Manas Kala$^\dag$, Anmol Rastogi$^*$, Kunal Dahiya$^\P$, Aya Hagishima$^\Phi$}\\
\IEEEauthorblockA{$^\Phi$ Interdisciplinary Graduate School of Engineering Sciences, Kyushu University, Fukuoka, Japan\\ $^\dag$ Mobile Computing Laboratory, Graduate School of Information Science \& Technology, Osaka University, Japan\\ $^*$ Indian Institute of Technology Hyderabad, India $^\P$ Indian Institute of Technology Delhi, India\\
Email: lala.betty.919@s.kyushu-u.ac.jp, manas\_kala@ist.osaka-u.ac.jp, ai19btech11021@iith.ac.in, kunalsdahiya@gmail.com, ayahagishima@kyudai.jp}}




\maketitle
\begin{abstract}
Indoor thermal comfort in smart buildings has a significant impact on the health and performance of occupants.
Consequently, machine learning (ML) is increasingly used to solve challenges related to indoor thermal comfort.
Temporal variability of thermal comfort perception is an important problem that regulates occupant well-being and energy consumption. However, in most ML-based thermal comfort studies, temporal aspects such as the time of day, circadian rhythm, and outdoor temperature are not considered. 
This work addresses these problems. It investigates the impact of circadian rhythm and outdoor temperature on the prediction accuracy and classification performance of ML models. 
The data is gathered through month-long field experiments carried out in 14 classrooms of 5 schools, involving 512 primary school students. Four thermal comfort metrics are considered as the outputs of Deep Neural Networks and Support Vector Machine models for the dataset.  
The effect of temporal variability on school children's comfort is shown through a ``time of day" analysis. Temporal variability in prediction accuracy is demonstrated (up to 80\%). Furthermore, we show that outdoor temperature (varying over time) positively impacts the prediction performance of thermal comfort models by up to 30\%. The importance of spatio-temporal context is demonstrated by contrasting micro-level (location specific) and macro-level (6 locations across a city) performance. The most important finding of this work is that a definitive improvement in prediction accuracy is shown with an increase in the time of day and sky illuminance, for multiple thermal comfort metrics.



\end{abstract}




\section{Introduction}
\label{sec:sample1}

Smart cities are envisioned as a complex mesh of cyber-physical systems that enable an intelligent and sustainable urban environment. Smart buildings and indoor spaces are the most essential components of this vision, as humans spend 90\% of a typical day in buildings and indoor spaces. Consequently, indoor environmental quality (IEQ), especially thermal comfort (TC), is vital for the occupants of a smart building. Indoor thermal comfort affects our physical health, emotional well-being, and productivity~\cite{[2]}. 


Machine Learning (ML) is increasingly being applied to model and predict the perceived thermal comfort of occupants. Compared to ML-based prediction models, the conventional techniques for TC estimation viz., Fanger's Predicted Mean Vote (PMV-PPD) model \cite{Fanger} and the adaptive model (ATC) \cite{Adaptive}, fail to offer high accuracy and generalization across spatio-temporal contexts \cite{ML_TC_REVIEW_1, ML_TC_REVIEW_2}.
This is due to the fact that TC perception is unique to an individual and highly subjective. 
ML algorithms can solve complex TC classification problems by learning multi-dimensional non-linear mappings between several environmental parameters and predict subjective preferences with high accuracy. 

 The sustainable smart city paradigm also has a comfort vs. sustainability trade-off that ML-based TC prediction can solve. Over 95\% of people in developing countries live in naturally ventilated (NV) buildings \cite{ventilation}. 
 Thus, a combination of IoT sensors and ML-based TC prediction is an effective solution for a sustainable yet smart indoor living~\cite{bettySmartsys}.

\subsection{Motivation and Research Problems}
Despite the recent focus on applications of ML, the problem of temporal variability in TC prediction is yet unexplored. 
In fact, the guidelines prescribed under the international TC standards, viz. ASHRAE 55 \cite{ANSIASHRAE2020} and EN 15251  
, assume TC perception to be constant throughout the day~\cite{ANSIASHRAE2020}. This can be attributed to decades-old TC studies conducted in controlled environments that found little impact of time of day on occupants' thermal preference \cite{timeold1, timeold2}. 

In sharp contrast, new research has shown a strong relationship between human TC perception and time of day \cite{timenew1_18, VelleitimeAnalysis}. Two crucial time-specific factors are highlighted. 
The first is the temporal variation in sky illuminance and light radiation that affects our circadian rhythm and, in turn, our perception of thermal comfort \cite{VelleitimeSurvey, VelleitimeAnalysis}. Second, the outdoor temperature at any given instant \cite{TimeTemp}. Moreover, the importance of temporal variability in TC perception is shown to optimize energy use and is also desirable for occupant health 
\cite{timehealthbenefits_21}. 
The new body of work on the impact of temporal variability is primarily empirical, and insightful conclusions are drawn through data analysis and statistical analysis \cite{VelleitimeAnalysis, TimeTemp}. 
However, the impact of temporal variability on the predictive modeling of subjective TC perception is yet lacking \cite{ML_TC_REVIEW_1, ML_TC_REVIEW_2}. Demonstrating the impact of temporal variation on TC prediction can motivate fine-tuning of the standards that will improve energy efficiency and occupant well-being in a smart built environment. In addition, given the spatial variability in TC perception and prediction in NV buildings, spatio-temporal analysis is also necessary~\cite{bettySmartsys}.
 
This work seeks to fill these gaps.
by demonstrating the impact of the temporal variation on TC prediction of occupants. In particular, it addresses three pertinent questions: (a) How time-specific should a TC prediction model be? (b) What is the impact of temporal variation in outdoor temperature on the performance of the TC prediction? (c) How significant is spatial variability (e.g., a specific building vs. a group of buildings) compared to temporal variability?

From the perspective of energy conservation and sustainability, these questions are highly relevant for NV buildings. First, the occupants of NV buildings are more vulnerable to temporal variability, as an HVAC unit does not thermally regulate the indoor spaces. Second, understanding temporal variability in TC perception further optimizes energy use~\cite{timeenergybenefits_20}. 

When it comes to occupants, children in NV buildings are the most vulnerable due to their low metabolic rates, limited cognitive abilities, and lack of thermal adaptability ~\cite{[59]}. Thus, students in NV classrooms are highly susceptible to the external environment that changes with time during the day. 
However, TC prediction studies for students in general and primary students in specific, are extremely rare \cite{ML_TC_REVIEW_1, ML_TC_REVIEW_2}. 

Therefore, this work considers an extremely challenging context that includes NV buildings and young primary school students through real-world field experiments.

\color{black}

\subsection{Contributions}
This work aims to solve the research problems concerning the temporal variability of TC perception and the consequent impact on the performance of TC prediction. Data is gathered from month-long field experiments conducted in 14 NV classrooms of 5 schools, involving 512 primary school students. ML models are created using Deep Neural Network (DNN) and Support Vector Machine (SVM) algorithms. 
The major contributions are described below:
\begin{enumerate}
     \item Three new features (Time of day, outdoor temperature, and satisfaction with clothing) and a new TC Metric (Thermal Satisfaction Level) are added to the baseline primary student dataset \cite{bettySmartsys, bettydeepcomfort}.
     \item Data Analysis: Time-specific exploratory 
     data distribution analysis 
     is performed to demonstrate the variation across six half-hour time slots. 
      \item Temporal Variability Analysis: The performance of DNN and SVM multi-class classification models for 4 TC metrics is evaluated 
      through a new ``time of day''  feature. 
     \item Relevance of Outdoor Temperature: DNN and SVM based prediction models are implemented for baseline primary student dataset and an upgraded dataset that has Outdoor Temperature as a new feature. 
     \item Spatio-Temporal Analysis: A comparative analysis of the impact of sky illuminance and outdoor temperature is performed for a specific location (single school) and at the city scale (all 6 schools).
     \item Insights and Validation: Unique inferences are drawn and results are validated through analysis of CIE Sky model illuminance data \cite{ISHRAE}.
 \end{enumerate}    
\begin{figure*}[htbp]
 \centering%
\begin{tabular}{cc}
\hspace*{\fill}
    \subfloat[Field Experiments in Different Time Slots] {\includegraphics[width=.35\linewidth]{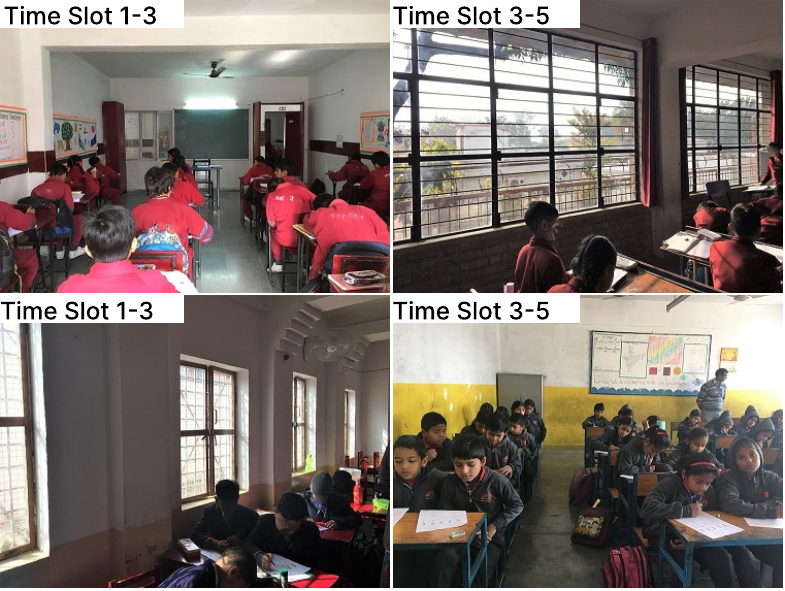} \label{fig:photos}}\hspace{0.3cm}%
	\subfloat[Simplified Questionnaire with Thermal Comfort Scales] {\includegraphics[width=.55\linewidth]{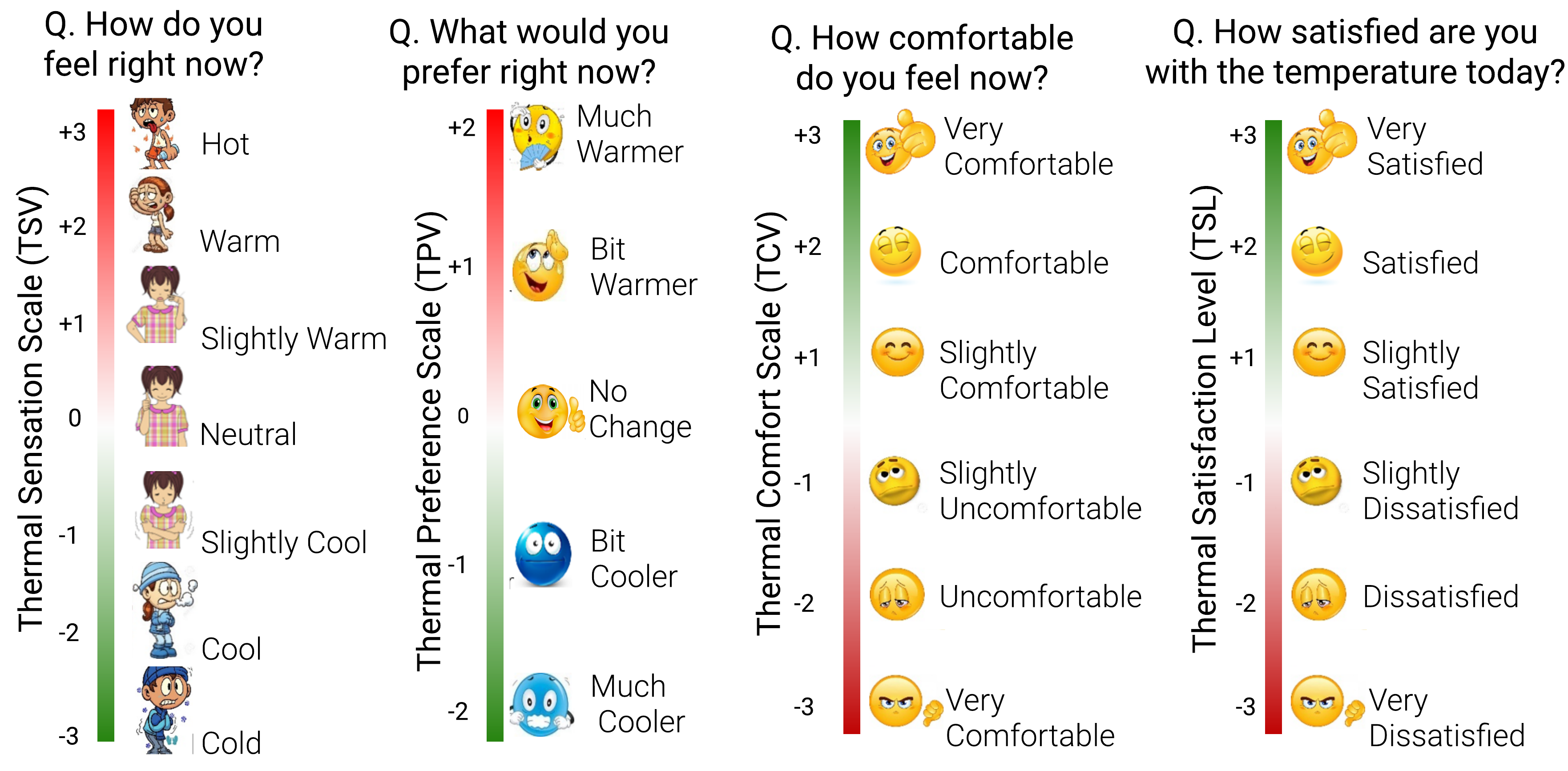}\label{fig:new_ques}}\\
\end{tabular}
  \caption{Field Experiments and Thermal Comfort Metrics}  
    \label{fig:expandques}
    \vspace*{-0.4cm}
\end{figure*}
\section{Literature Review and Relevant Aspects}
Thermal comfort of an individual is highly subjective 
and has multiple dimensions, such as sensation, preference, level of comfort, and level of satisfaction, etc. These dimensions are quantified through TC metrics such as the Thermal Sensation Vote (TSV), Thermal Preference Vote (TPV), Thermal Comfort Vote (TCV), Thermal Satisfaction Level (TSL), etc. ML algorithms use the TC metrics viz., TSV, TPV, and TCV, as outputs (labels), to accurately predict thermal comfort of the occupants \cite{ML_TC_REVIEW_1, ML_TC_REVIEW_2}. Any variability in TC perception of occupants will cause variation in TC metrics, which in turn is likely to influence the performance of TC prediction models. 

Thus, temporal variability in the performance of an ML-based TC prediction model needs to be studied. Surprisingly, a majority of older TC studies conducted between 1960-1990, have concluded that the time of day is insignificant~\cite{timeold2, timeold1, timeold3, Fanger}. However, some studies also presented results to the contrary. For example, noticeable impact of time of day on ambient temperature preferences was also reported when contrasting morning with afternoon \cite{timeold4_yes}.
\par\textbf{{State-of-the-art:}}
In the past 5 years, significant new evidence has been brought to light that demonstrates that perception of thermal comfort is dynamic and undergoes temporal variation~\cite{VelleitimeAnalysis, timenew1_18}. A recent study on human physiology demonstrated that the Basal Metabolic Rate (BMR), i.e., the human resting metabolic rate, varies with the circadian rhythm of the human body \cite{basal_11}.
However, BMR is only one of the variables that varies in a day. 
In an exhaustive review of temporal variation studies in indoor thermal comfort, Vellei et al. focused on thermoregulatory behaviors and their impact on circadian rhythms~\cite{VelleitimeSurvey}. 


The most concrete evidence of temporal variation is presented in \cite{VelleitimeAnalysis} through statistical analysis of SCATs (Smart Controls and Thermal Comfort) and ASHRAE-I databases~\cite{db1}. The study demonstrates the difference in thermal preference at different times and hypothesizes that it is caused by human circadian rhythm. Similarly, time of day and light exposure is shown to affect human TC perception~\cite{daylightTC}. For example, the TSV and TPV may differ significantly at different times during a day~\cite{TimeTemp}. Furthermore, ``time of day" may impact the occupants emotional response, that affects their thermal response~\cite{ko2020impact}. 
Understanding the temporal variability in TC perception can help optimize energy efficiency of buildings while ensuring occupant comfort~ \cite{timeenergybenefits_20}. Equally importantly, due to the thermoregulatory and physiological implications of time-varying thermal comfort perception, it is also highly desirable for occupant health \cite{timehealthbenefits_21}.

Several recent studies have highlighted the absence of a ``time of day" variable in most TC studies and strongly recommends its inclusion\cite{VelleitimeSurvey, daylightTC, VelleitimeAnalysis, TimeTemp}.
This holds even more true for ML-based TC prediction, where an important feature like ``time of day" can significantly enhance prediction performance. The two most important aspects that have emerged from recent findings are discussed ahead, along with the open problems.

\par\textbf{{Temporal Variation and Circadian Clock:}}
The human circadian clock is regulated and synchronized by light exposure from solar radiation and illuminance~\cite{lightTC_23,VelleitimeSurvey}. Therefore, light exposure directly impacts occupant TC perception and determines the preferred ambient temperature~\cite{daylightTC}. The external photic input primarily depends on two factors, viz., the intensity of solar radiation and the percentage and opacity of cloud cover. The resulting diffused radiation is the light exposure that regulates biological circadian rhythm, and in turn, occupant thermal comfort perception and health. Diffused radiation is highly temporal and varies as the day progresses. Apart from the time of day, it is also dependent on weather and seasons. Further, in NV buildings, the impact of external light exposure is heightened because doors and/or windows, etc., are usually opened to facilitate cross ventilation. Thus temporal variability due to light exposure is a greater concern in NV buildings as compared to HVAC buildings that may block out all external light exposure (by using blinds or curtains) and rely entirely on artificial indoor lighting.

\par\textbf{{Temporal Variation and Outdoor Temperature:}}
The second most important feature with respect to temporal variation is the outdoor temperature. The importance of outdoor temperature in estimating thermal sensation and comfort is highlighted along with daylight in \cite{daylightTC}. Further, outdoor temperature is one of the temporal factors affecting the feature relationship between TSV and PMV, learned through linear regression \cite{VelleitimeAnalysis}. Likewise, variation in outdoor temperature during a day affects the thermal perception of students in a classroom in \cite{TimeTemp}.
\par\textbf{{Open Problems:}}
The primary unaddressed challenge in temporal variability analysis is to demonstrate how the ability of conventional (e.g., SVM) and advanced ML algorithms (e.g., DNN) to predict occupant TC varies with time. This can be done  by considering a ``time of day" feature in the model. For example, does the ML model performance improve with greater light exposures? Further, the impact of outdoor temperature on time-specific predictive modeling needs to be demonstrated. A spatio-temporal analysis will also help understand localized impact of time of day and outdoor temperature on TC prediction. The predictive modeling and analysis presented in the sections ahead extend the existing knowledge base by offering crucial insights to these important questions.
\color{black}

\begin{figure}[H]
    \includegraphics[width=\linewidth]{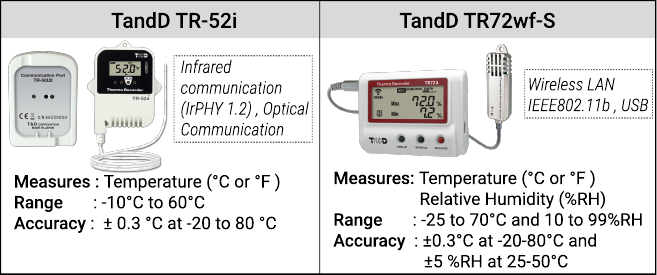}
    \captionof{figure}{IoT Sensors used in Measurements}
    \label{fig:sensoriot}
    \vspace{-0.3cm}
\end{figure}
\begin{figure*}[ht!]
 \centering%
\begin{tabular}{cc}
\hspace*{\fill}
    \subfloat[Thermal Sensation Distribution] {\includegraphics[width=.33\linewidth]{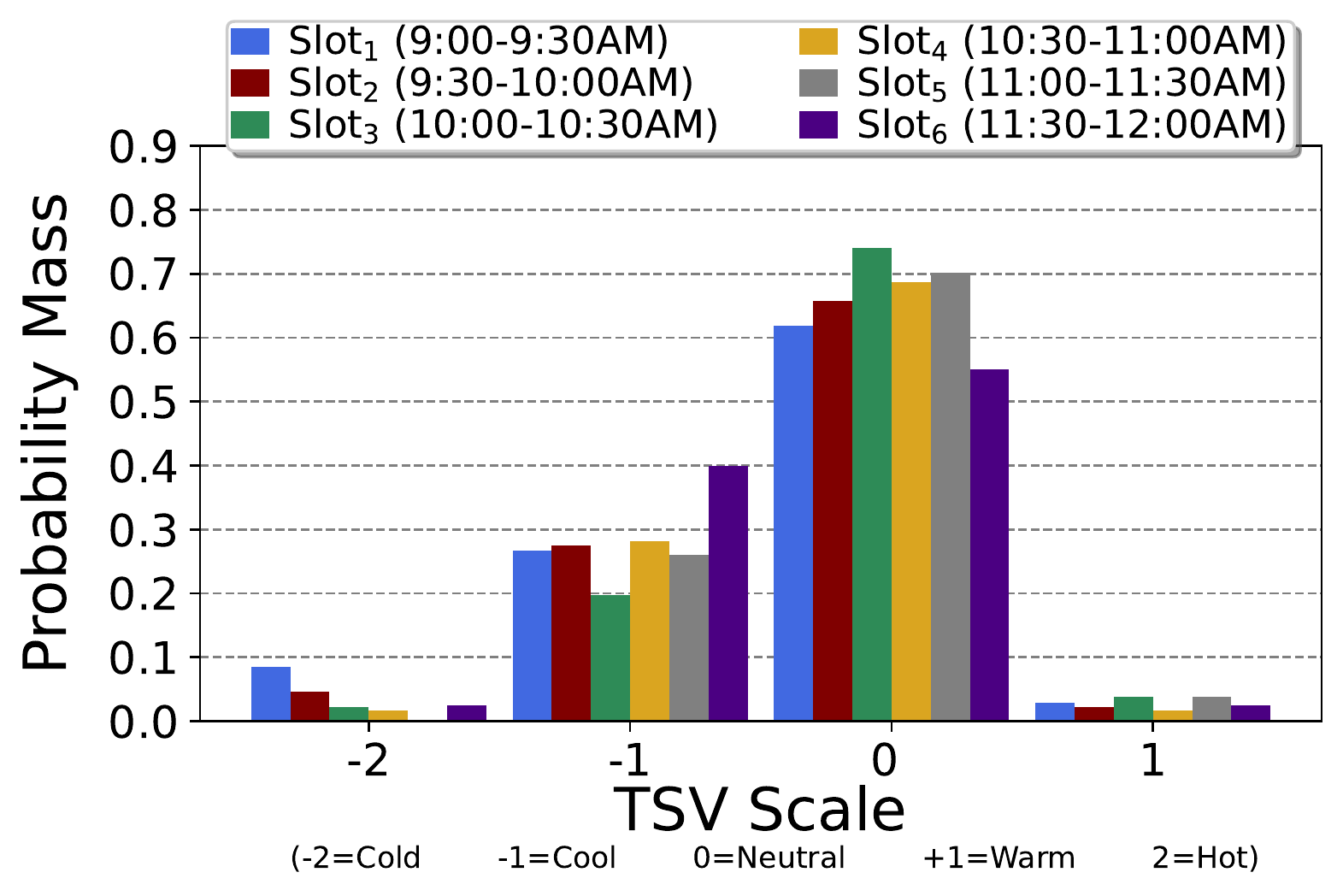}}\hspace{0.1cm}
    \subfloat[Thermal Preference Distribution] {\includegraphics[width=.28\linewidth]{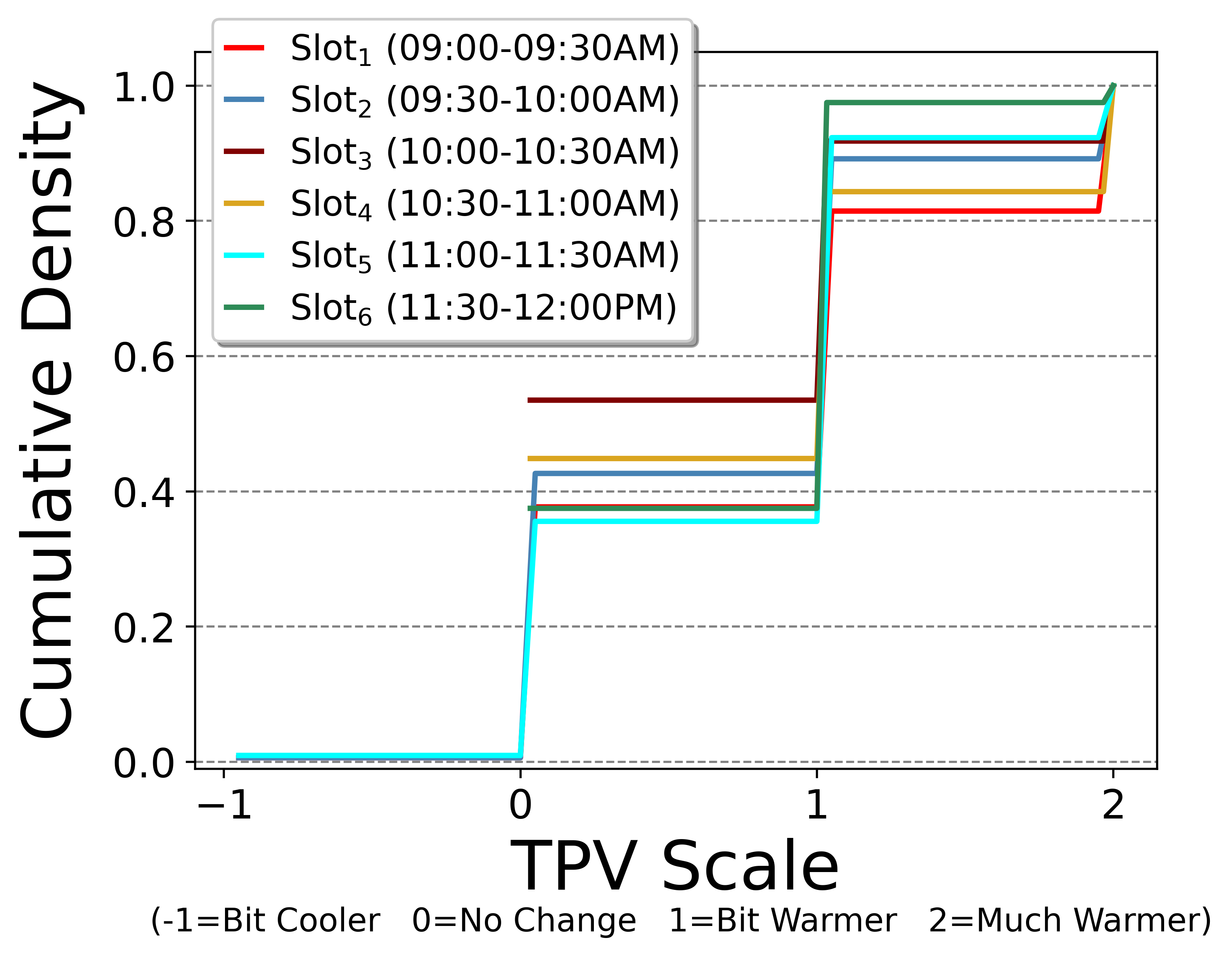}\hfill}
      \subfloat[Thermal Comfort Distribution] { \includegraphics[width=.35\linewidth]{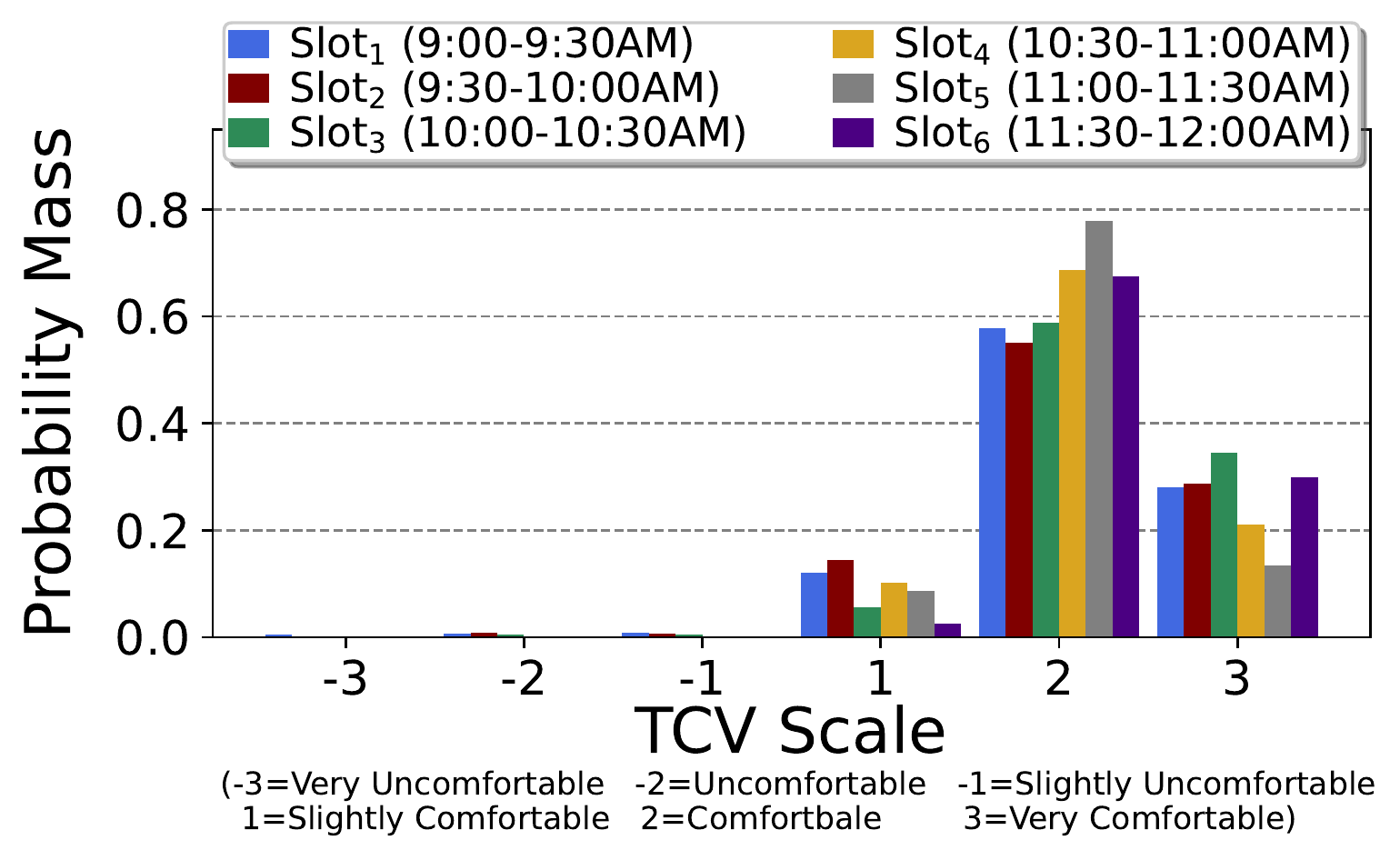}}\hspace*{0.2cm}

\end{tabular}
  \caption{Distributions of Output Metrics and Time of Surveys}  
    \label{fig:scales}
    \vspace*{-0.4cm}
\end{figure*}

\section{Field Experiments and Data Gathering}
The field experiments were conducted in 5 schools of Dehradun city, India, in January, the coldest month of the year. The city is located in the foothills of the Himalayas, and enjoys a composite climate~\cite{bettyIndia}.

The experiments were conducted in 14 NV classrooms to investigate the thermal comfort perceptions and preferences of 512 unique primary school students. The 5 schools that participated in the study are, St. Thomas School, Grace Academy, Cambrian Hall, Kendriya Vidhyalaya, and Jaswant Model School. In this work, numbers 1 to 5 are randomly assigned to each school to ensure anonymity and confidentiality. The students belonged to grades 3 to 5, and their ages ranged from 6 to 13 years. A total of 2039 samples were collected from field experiments that lasted 28 days. 

A detailed description of the questionnaire and field-experiment methodology is available in \cite{bettydeepcomfort, bettySmartsys, bettySust}. The new discussion and data analytics presented in the next subsections is specific to temporal variation. 

\subsection{Field Experiments and Data Gathering}
Subjective assessment responses of "right here right now" were collected from students to explore their thermal perceptions in classrooms. Figure~\ref{fig:photos} shows students completing the questionnaire survey during the experiments. The questionnaire was designed with illustrations and simple vocabulary, and was explained in detail before every experiment. Subjective rating TC scales like TSV, TPV, TCV and TSL were simplified for students' better understanding, as illustrated in Figure~\ref{fig:new_ques}. The clothing values were calculated by adding their individual clo values collected from the questionnaire. Personal information included age, grade, and gender.
\begin{table}[H]
\footnotesize
\begin{centering}
{
 \captionof{table}{Statistical Details of Field Experiment Time Slots}
  \label{table:timeslotstats}
  \begin{tblr}{
      colspec={p{1.9cm}p{0.4cm}p{0.6cm}p{0.65cm}p{0.6cm}p{0.65cm}p{0.8cm}},
      row{1}={font=\bfseries, bg=gray!10},  
      row{even}={bg=blue!10},
    }
    \toprule
     Timing&Time  Slot & Schools & Grades & Age (Avg.) & Clo Values& Samples \\
   \toprule
    09:00-09:30AM&1&2,3,4   &3,4   &8.8  &1.43 &647\\
    09:30-10:00AM&2&1,2,3,4 &3,4,5 &9.9  &1.39 &647\\
    10:00-10:30AM&3&1,3,4,5 &3,4,5 &9.5  &1.35 &415\\
    10:30-11:00AM&4&1,4,5   &3,4,5 &9.6  &1.36 &185\\
    11:00-11:30AM&5&1,4     &4,5   &10.1 &1.24 &104\\
    11:30-12:00PM&6&1       &5     &10.3 &1.24 &40\\
    \bottomrule
  \end{tblr}
}
\end{centering}
  \vspace{-0.3cm}
\end{table}

\par\textbf{Timing Specific Methodology:} Objective physical measurements were performed simultaneously while students filled in the questionnaire. Two IoT sensors were used, viz., \mbox{TandD TR72wf-S}~(measures the indoor temperature and relative humidity) and \mbox{TandD TR72wf-S}~(measures the outdoor temperature). The technical specifications of the sensors are provided in Figure~\ref{fig:sensoriot}.  

The experiments were conducted on consecutive days (4-5 days/school) during school hours, usually targeting the earlier hours when it was coldest and sky illuminance was low. The study considered 6 time slots, each of half-hour duration, from 9:00 AM to 12:00 PM, represented as Slot$_{i}$, where $i\in \{1\ldots6\}$. The strategy ensured a progression from (a) low to maximum outdoor temperatures and (b) low to maximum light exposure, on a winter day, serving the objectives of this study.
The statistical details for each time slot are summarized in Table~\ref{table:timeslotstats}. 


\par\textbf{Updated Feature Set and TC Metrics (Labels):} 
The original dataset collected from the field experiments contained 13 features and 7 subjective labels (TC metrics)~\cite{bettySmartsys, bettydeepcomfort}. These include the objective input variables such as maximum and minimum daily temperatures, indoor temperature and relative humidity, clothing values, metabolic rates, etc., and  psycho-social features like age, gender, and grade.

This work considers 3 new features as compared to our original primary student dataset~\cite{bettySmartsys, bettydeepcomfort, bettySust}. The new features are (i) Running mean outdoor temperature for the duration of the experiments collected through TandD TR-52i (Figure~\ref{fig:sensoriot}),  (ii) A categorical Time of Day feature (``Starttime"), that marks the start-time of each experiment and student response, and (iii) A subjective feature to capture students' satisfaction with clothes (wearing excess, satisfied and wearing less). 
Likewise, Thermal Satisfaction Level (TSL) responses are gathered from students and is a new TC Metric (label) used in the comfort prediction modeling in this work. TSL is rarely used in the estimation and prediction of thermal comfort. We believe that TSL is particularly useful for students who have limited cognition and can express ``satisfaction" more easily than ``preference" or ``comfort."
\footnote{Please note that 
The proposed analysis could not be performed on the standard ASHRAE-II~\cite{db2} dataset, as both (a) timing data and (b) sufficient data for children are not available.}

 \begin{figure}[H]
\centering 
    \includegraphics[width=0.8\linewidth]{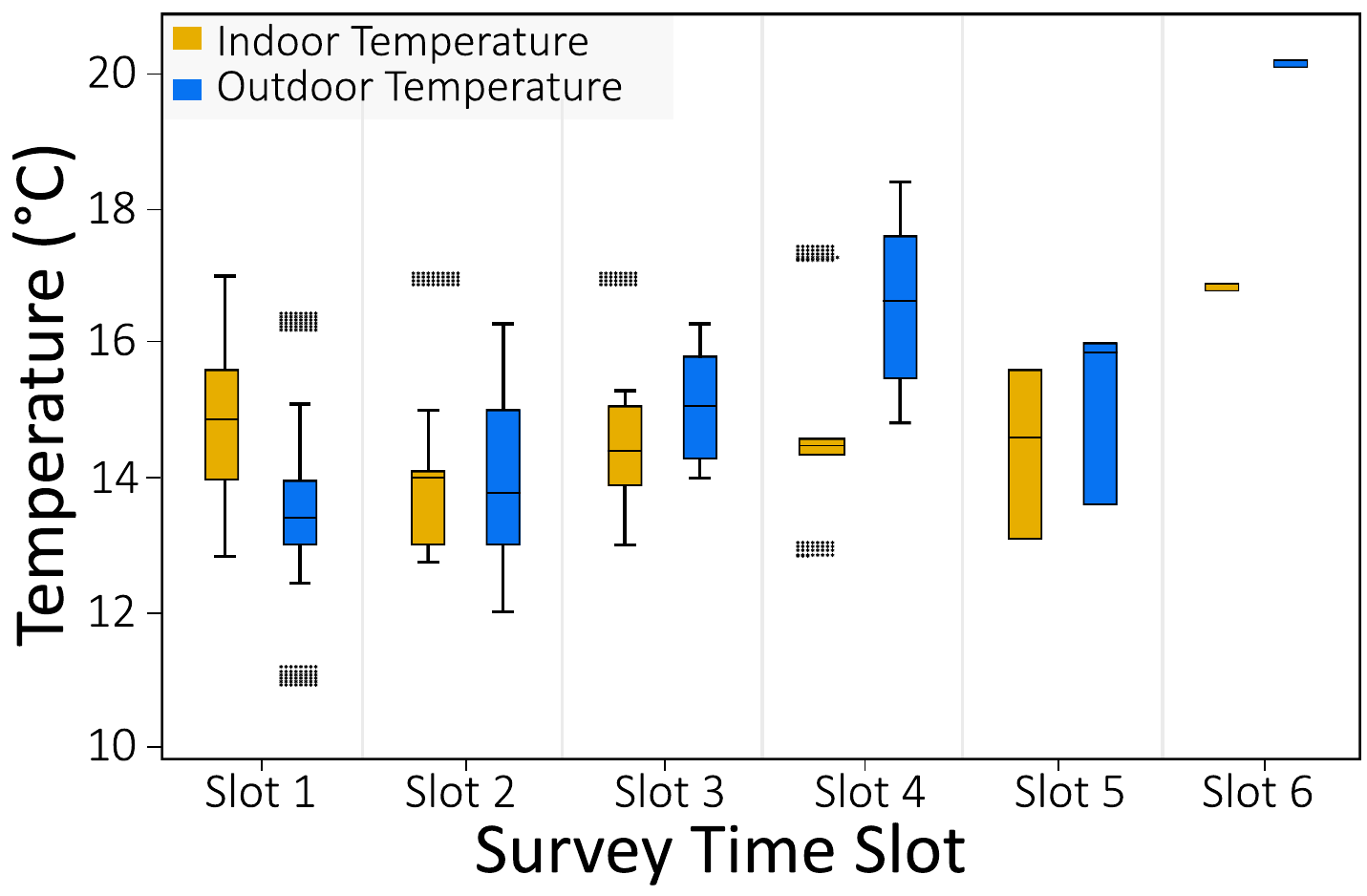}
\caption{Indoor and Outdoor Temperature Distribution}
\label{fig:tsvs}
\vspace{-0.3cm}
\end{figure}
\subsection{Exploratory Data Analysis}

The subjective responses of students, presented in Figure~\ref{fig:scales}(a), shows that the proportion of students feeling ``Cold'' (TSV=-2), reduces as the day progresses, i.e., $i$ in Slot$_{i}$ increases. Although this can be attributed in part to ambient indoor temperatures, it is noteworthy that median indoor temperature hovers around 15°C for all time slots, which can be observed in Figure~\ref{fig:tsvs}. Secondly, the median outdoor temperature increases till TimeSlot\textsubscript{4}, and then decreases. A corresponding increase in students feeling ``Cool'' (TSV=-1) can be noticed in  TimeSlot\textsubscript{4} and TimeSlot\textsubscript{5}. Thus, similar to the findings in \cite{TimeTemp}, outdoor temperature seems to be affecting thermal sensation of students. It's impact on TSV prediction performance remains to be seen.

The cumulative distribution of preference responses is illustrated in Figure~\ref{fig:scales}(b). The percentage of students who prefer a warmer classroom (TPV= 1 or 2) should ideally reduce in later time slots. However, this percentage is higher in TimeSlot\textsubscript{4}, than in TimeSlot\textsubscript{2} and TimeSlot\textsubscript{3}. This cannot be explained by the average indoor temperature which constantly increases from 14.00°C in TimeSlot\textsubscript{2} to 16.90°C in TimeSlot\textsubscript{6}. It can also not be explained on the basis of the outdoor temperature, which increases from 13.84°C in TimeSlot\textsubscript{2} to 16.62°C in TimeSlot\textsubscript{4}. Variation in light exposure could be a possible reason for this unexpected variation in TC perception. In the analysis of the sky illumination and impact of circadian rhythm presented ahead, concrete evidence for this argument will be presented.

Thus, prima facie, both light exposure or circadian rhythms and variation in outdoor temperature seem to have an impact on subjective TC perceptions of children.
Further, the complexity of predicting thermal comfort of children in a naturally ventilated building becomes apparent by observing the distributions of TSV, TPV, and TCV in 
Figure~\ref{fig:scales} 
The thermal sensation and thermal preference responses of children do not conform to their assessment of their thermal comfort level. Most of the students indicate that the classroom environment is comfortable. However, their TSV and TPV responses indicate a sense of discomfort and need for change. This presence of \textit{illogical votes} in the data can be attributed to the limited cognition of children. This problem can severely impact accuracy as data bias confuses the classifier~\cite{bettySmartsys}. Unfortunately, due to the limited number of ML-based prediction works focused on children, little emphasis has been given to this problem.
Thus, selecting the ideal time for thermal comfort surveys and measurements is challenging. 

Our study addresses this problem by considering a wider range of time-slots with varying outdoor and indoor temperatures. This applies especially to primary school students, given the additional challenges that their subjective responses present to the task of ML based predictive modeling. 


\color{black}

\begin{figure}[H]
\centering 
    \includegraphics[width=0.9\linewidth]{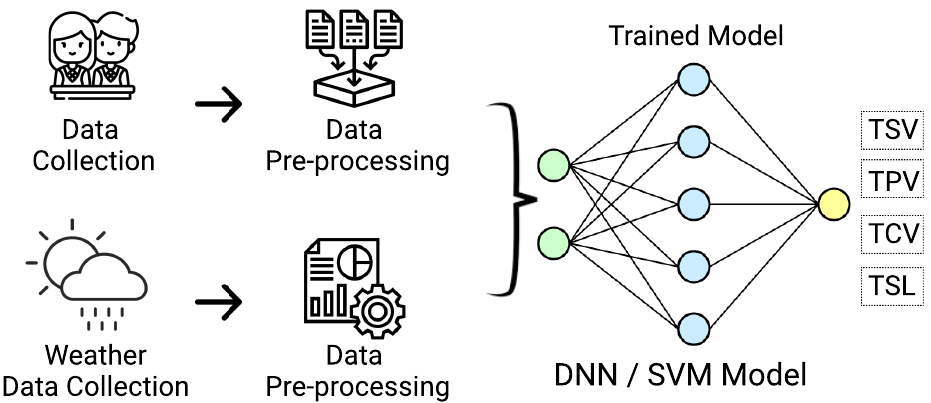}
\caption{System Design Overview}
\label{fig:system}
    \vspace{-0.3cm}
\end{figure}

\begin{figure*}[h]
 \centering%
\begin{tabular}{cc}
      \subfloat[ Solar Illuminance During Experiments] { \includegraphics[height=4.5cm, width=.36\linewidth]{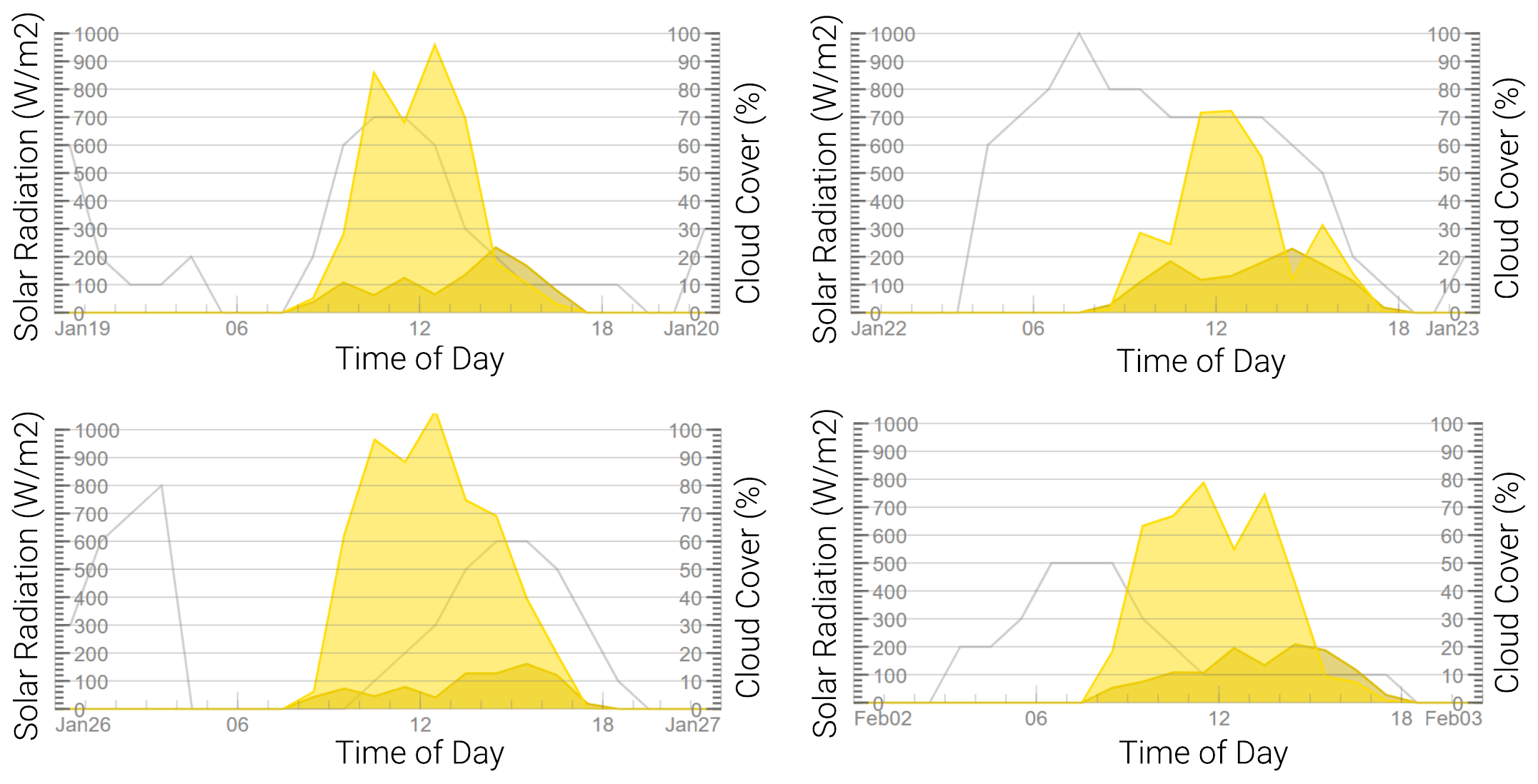}}
    \subfloat[Time-Slot Specific Accuracy for DNN] {\includegraphics[width=.32\linewidth]{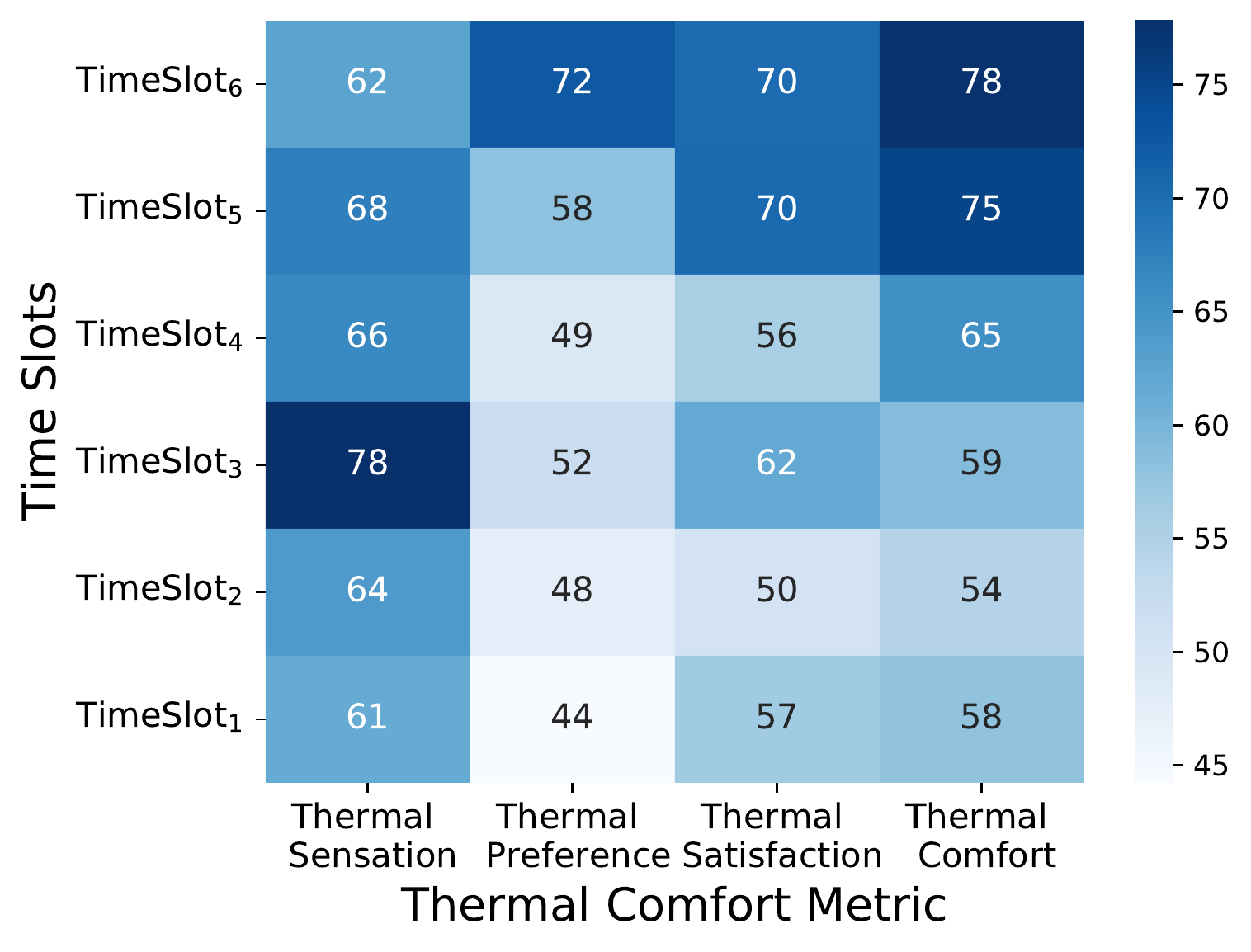}}
	\subfloat[Location Specific (School\textsubscript{1}) Accuracy for DNN] {\includegraphics[width=.32\linewidth]{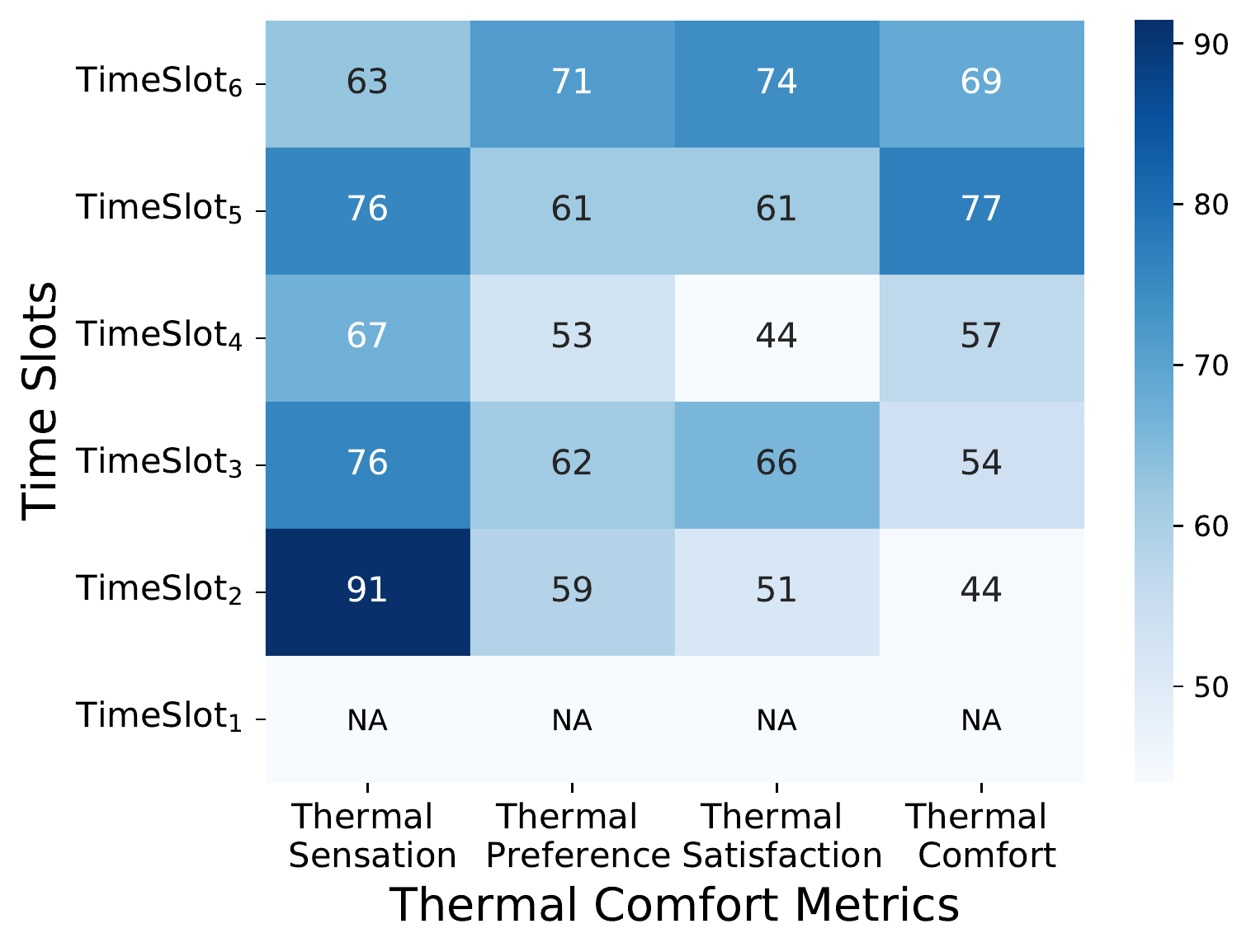}}

\end{tabular}
  \caption{Impact of Solar Illuminance and Light Exposure (Circadian Rhythm) on Prediction Accuracy of DNN}  
    \label{fig:circadian}
\end{figure*}

\section{System Implementation of Prediction Models} \label{sec:system}

To assess the impact of temporal variability, Deep Neural Network (DNN) and Support Vector Machine (SVM) based multi-class classification models are implemented for the four subjective TC metrics, viz., TSV (sensation), TPV (preference), TSL (satisfaction), and TCV (Comfort). The system design is illustrated in Figure~\ref{fig:system}. Experiments are performed for the following test-scenarios.
\par\textbf{Objectives:}  (i)~\textit{Temporal Variability and Circadian Rhythms}:  Investigate the variation in the classification performance of DNN and SVM models for (a) All time slots and (b) Each specific time slot. Here, DNN \& SVM models were trained on the entire dataset, and then tested on samples from all time slots, and individual time slots, respectively. (ii)~\textit{Temporal Variability and Outdoor Temperature}: Investigate the impact of outdoor temperature by comparing prediction model performance with and without the new feature of \textit{running mean outdoor temperature} introduced in this work. (iii)~\textit{Spatio-Temporal Variability}: Investigate the micro space-specific (i.e., single school) temporal variability for the first two test-scenarios, and compare it with city-scale (i.e., all 5 schools) TC prediction performance. 
\par\textbf{Problem Formulation:}
Let the primary student dataset be defined by $\mathcal{D}: \{(\v x_i, \v y_i)\}_{i=1}^{S}$, $S$ is number of samples, $\mathcal{C}$ is number of TC Metric classes. Further, $\v x_i \in \mathbb{R}^d$ represents the feature (input) vector, and $\v y_i \in \{0, 1\}^{\mathcal{C}}$ represents the label (output) vector for $i^{th}$ point. 

The SVM and DNN formulations, technical specifications, and evaluation strategies are as follows.

\par\textbf{Support Vector Machines~(SVM):} 

The SVM model tries to solve the following optimization problem for the TC Metrics in the primary student dataset as a binary classification:
\begin{align} \label{svm_eqn}
    \min_{\v w, b} \frac{1}{2} \v w^{\top}\v w + C \sum_{i=1}^{N} {\zeta_i} \\  \mbox{s.t.      } y_i(\v w^{\top} \phi(\v x_i) + b) &\ge 1 - \zeta_i, \forall i \nonumber
    \zeta_i &\ge 0, \forall i \nonumber
\end{align}
where, $y_i$ is +1 if the label is assigned to $i^{th}$ instance and -1 otherwise. Here $\v w \in \mathbb{R}^d$ denotes the weight-vector and the bias is denoted by $b$. Eq.~\ref{svm_eqn} is solved using the libsvm algorithm~\cite{libsvm}. Further, the hyperparameter $C$ is set through cross-validation.  $C$ controls the L2 regularization term, which avoids overfitting, and thus enhances the SVM classifier's generalization ability on the samples not seen by the model. 
 TC prediction may require a non-linear fit. Thus, the kernel function $\phi$ learns non-linear class-boundaries through the dot product of two vectors ($\v u$, $\v v$) in arbitrarily large spaces ($\phi(\v u)^{\top} \phi(\v v)$) without explicitly projecting the vectors into high dimensional space ($\phi(\v u)$ or $\phi(\v v)$). Experiments are run for four kernels viz., Linear, Radial Bias Function, and Polynomial kernels order~2 and order~3. Most importantly, prediction of the four TC Metrics requires a multi-class classifier. Thus the binary SVM classifier in Eq.~\ref{svm_eqn} is transformed to a multi-class classifier through the ``One vs. Rest" strategy.

\par\textbf{Neural Networks~(NN):} Neural networks jointly learn a sequence of non-linear transformation as well as the classifier. In particular, the aim is to solve the following problem:
\begin{align} \label{nn}
    \min_{\v \Phi, \v \{\v w_j, b_j\}_{j=1}^{\mathcal{C}}} \frac{\alpha}{2} \sum_{j=1}^{\mathcal{C}} \v w_j^{\top} \v w_j + \sum_{i=1}^{N} \sum_{j=1}^{\mathcal{C}} - \v y_{ij} \log(\hat{y}_{ij})
\end{align}
where, $\v \Phi$ denotes the parameters of function $\phi$, $\alpha$ is the hyper-parameter to control regularization, and $\hat{y}_{ij} = \frac{e^{\v w_j^{\top} \phi(\v x_i)}}{\sum_{k=1}^{\mathcal{C}}e^{\v w_k^{\top} \phi(\v x_i)}}$ is the prediction probability of class $j$ for $i^{th}$ instance. Moreover, $\phi$ is a non-linear transformation learned via a sequence of linear transformations followed by a non-linearity. The Adam optimizer~\cite{Adam} was used to solve~\ref{nn}. We experimented with various configurations with $\{5, 10, 15, 20, 25, 50\}$ neurons, up to 2 hidden layers (in addition to input and output layers), and \{ReLU, Sigmoid, Tanh\} non-linearities. Since 13 or 14 features are considered, Overfitting was avoided by not using a large number of neurons and very deep neural architectures.

It is noteworthy that the primary student data has an inherent class imbalance in all TC Metrics (Figure~\ref{fig:scales}). To address this challenge, a weighted classification strategy is considered such that
$\beta_{+} = \frac{N}{\sum_{i=1}^{N}\mathbb{I}(y_{i}=1)}$ and $\beta_{-} = \frac{N}{\sum_{i=1}^{N}\mathbb{I}(y_{i}=-1)}$ are the weights assigned to the positive and negative classes, respectively. 

\begin{figure*}[htbp]
 \centering%
\begin{tabular}{cc}
    {\includegraphics[height = 3.5cm, width=.03\linewidth]{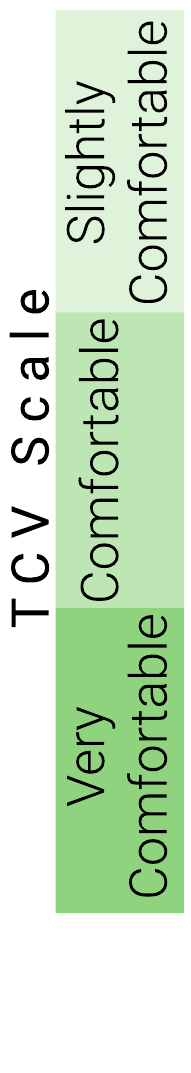}}
    \subfloat[All Time Slots (Acc=46, F1=49)] {\includegraphics[width=.23\linewidth]{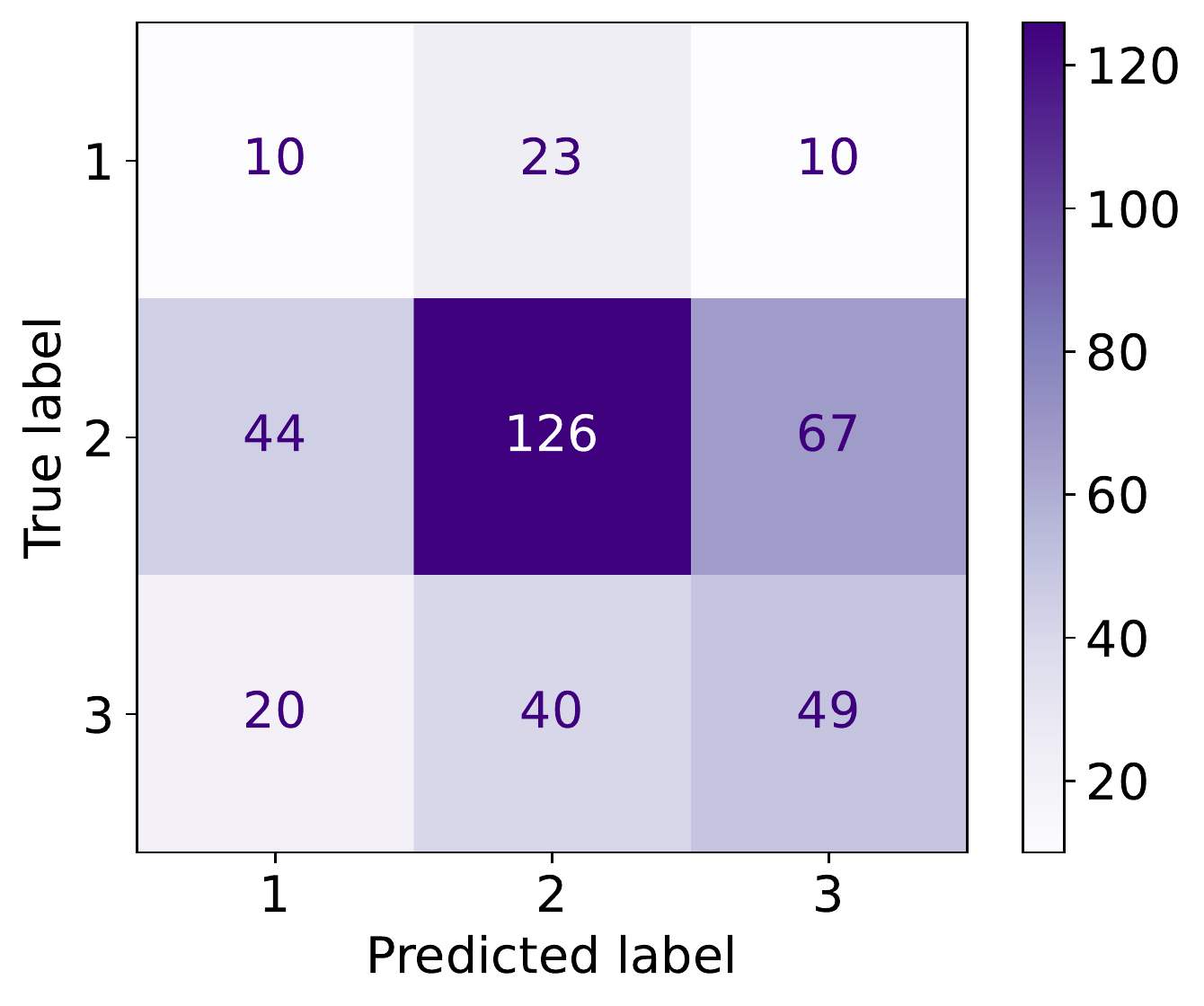}}
	\subfloat[TimeSlot\textsubscript{2} (Acc=40, F1=42)] {\includegraphics[width=.23\linewidth]{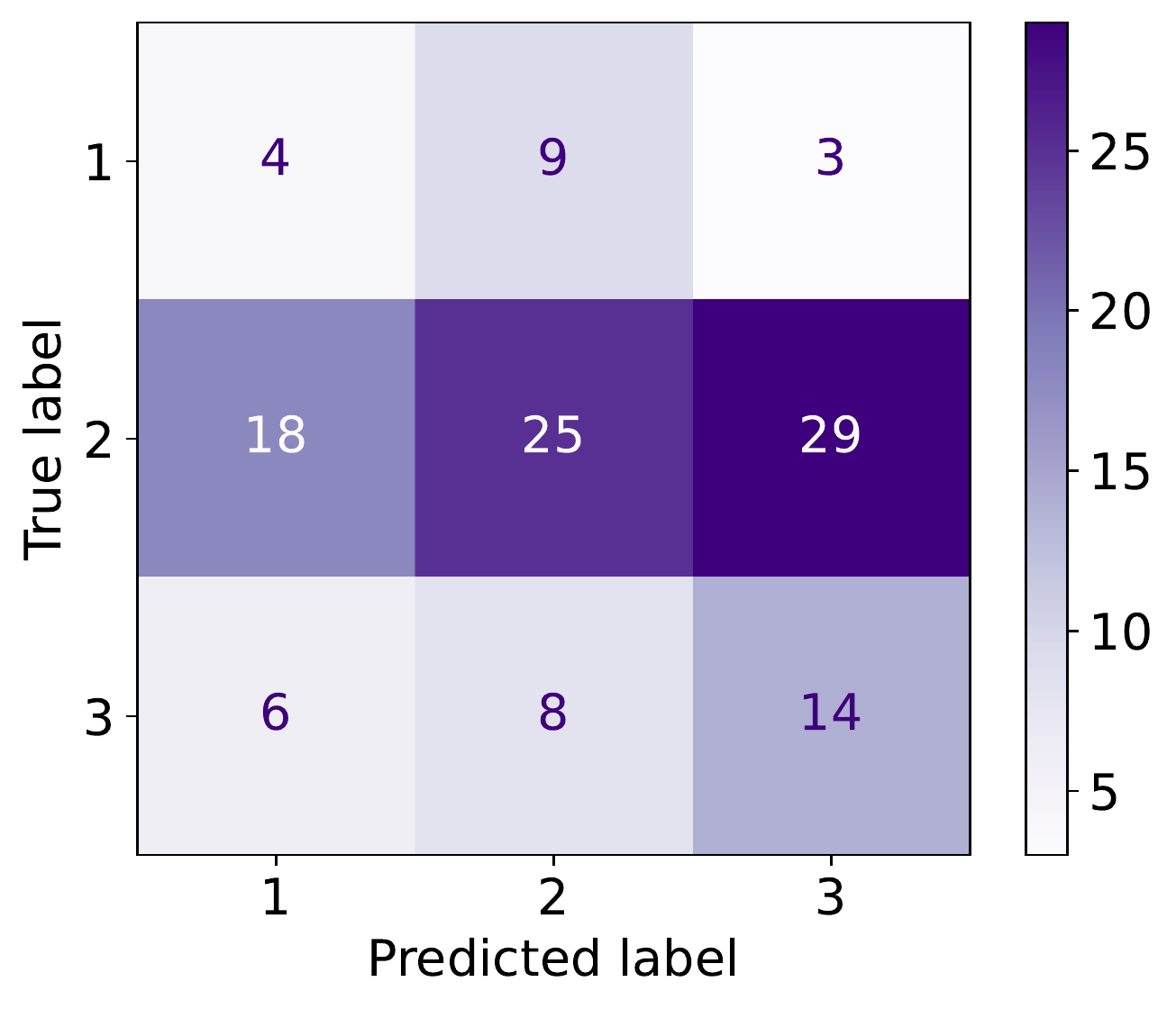}}
      \subfloat[TimeSlot\textsubscript{4} (Acc=67, F1=66)] { \includegraphics[width=.23\linewidth]{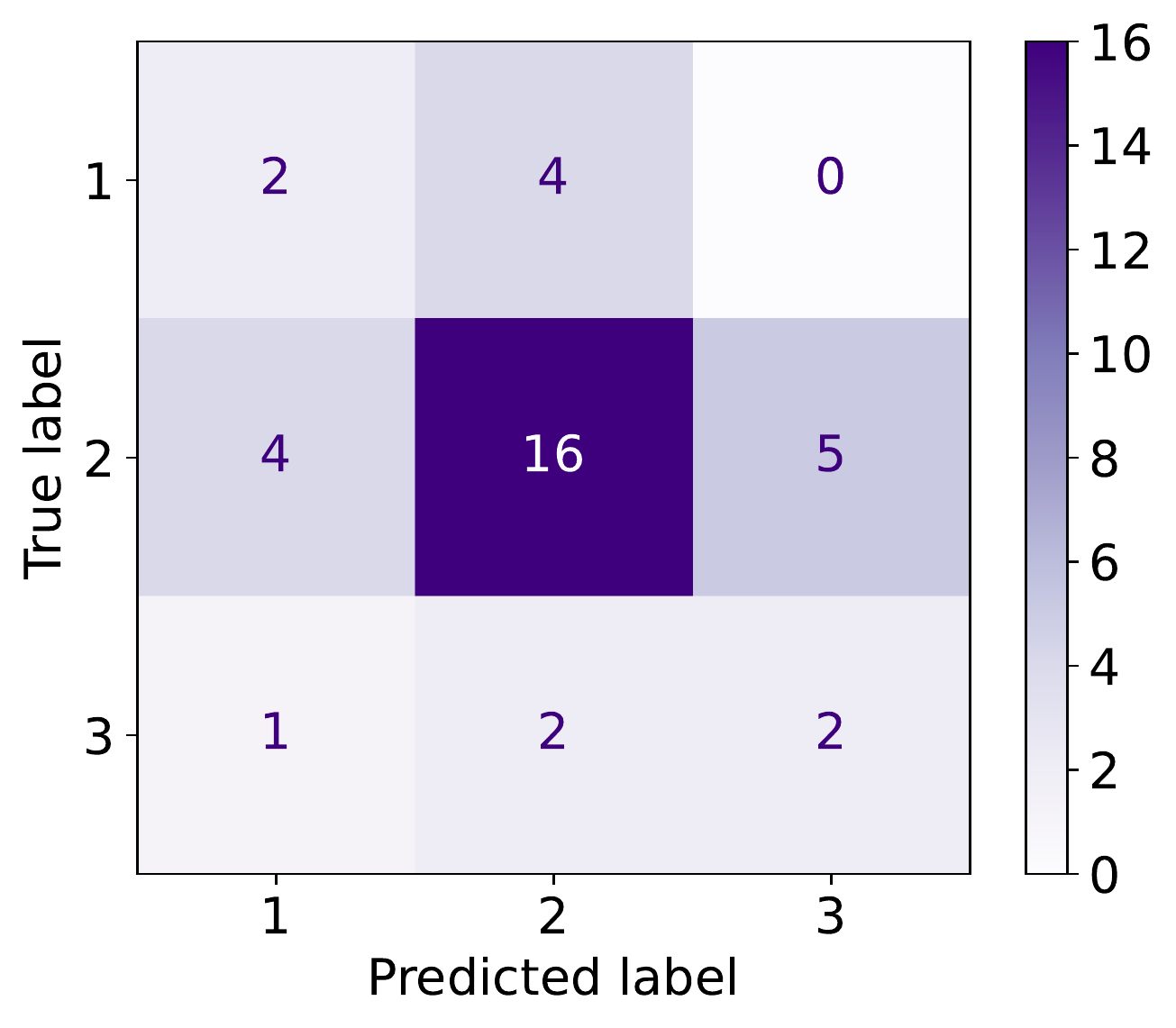}}
      \subfloat[TimeSlot\textsubscript{5} (Acc=80, F1=77)] { \includegraphics[width=.23\linewidth]{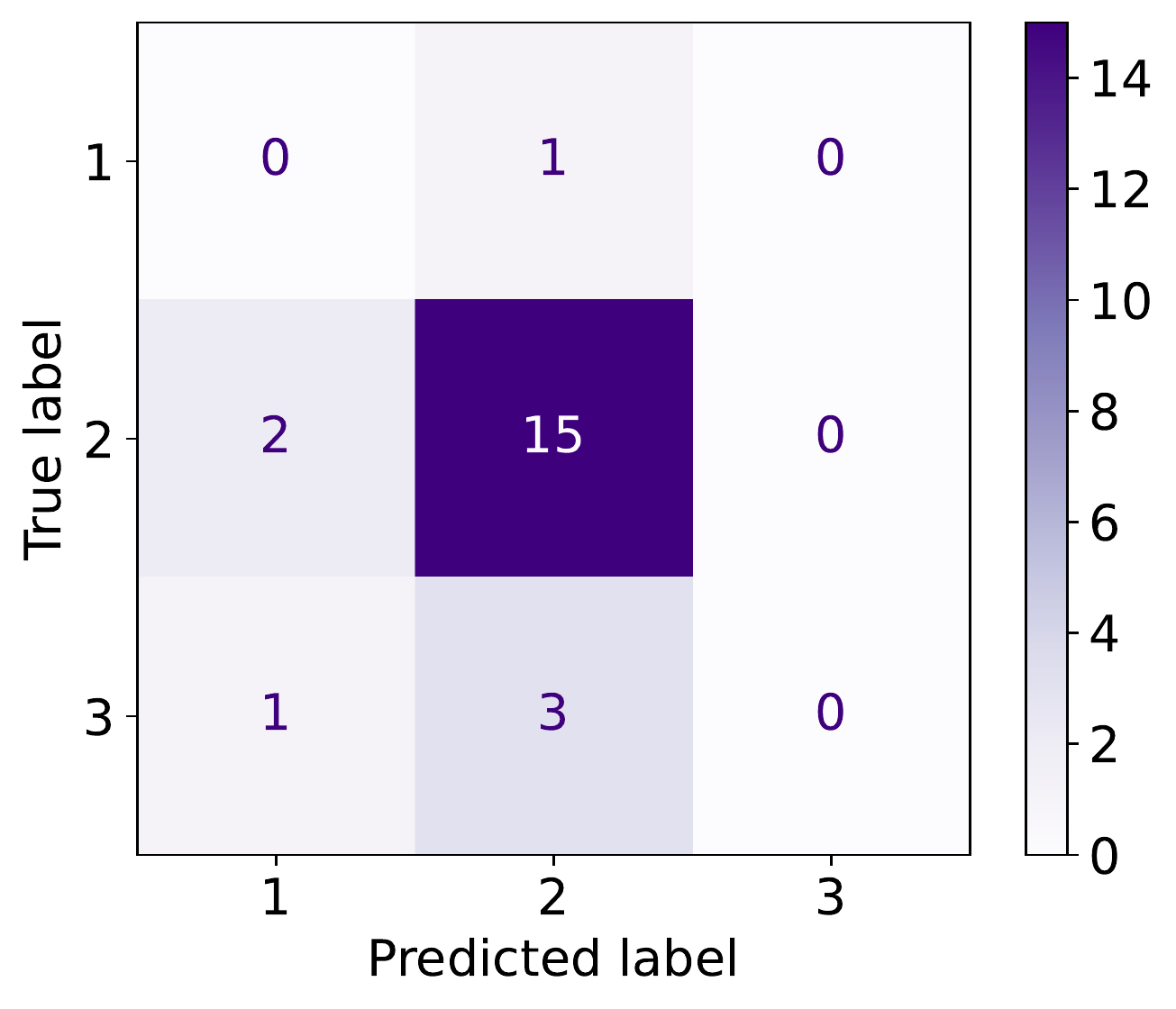}}
\end{tabular}
  \caption{Temporal Specificity of Multi-class Classification Performance of DNN for TCV}  
    \label{fig:classification}
    \vspace*{-0.4cm}
\end{figure*}

\begin{figure*}[htbp]
 \centering%
\begin{tabular}{cc}
\hspace*{\fill}
    \subfloat[All Time Slots Combined] {\includegraphics[width=.33\linewidth]{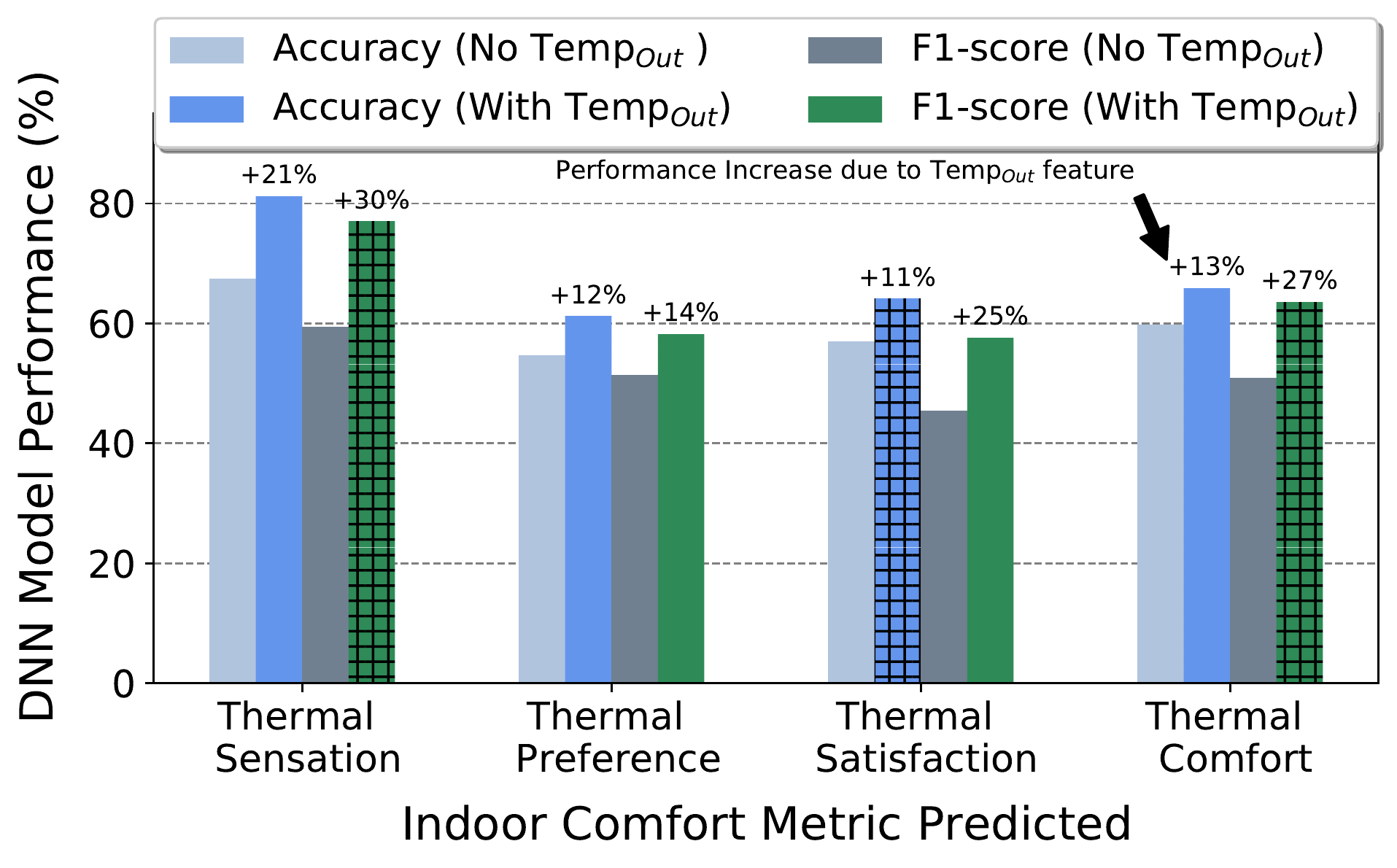}}\hspace{0.1cm}%
	\subfloat[Time Slot Specific] {\includegraphics[width=.33\linewidth]{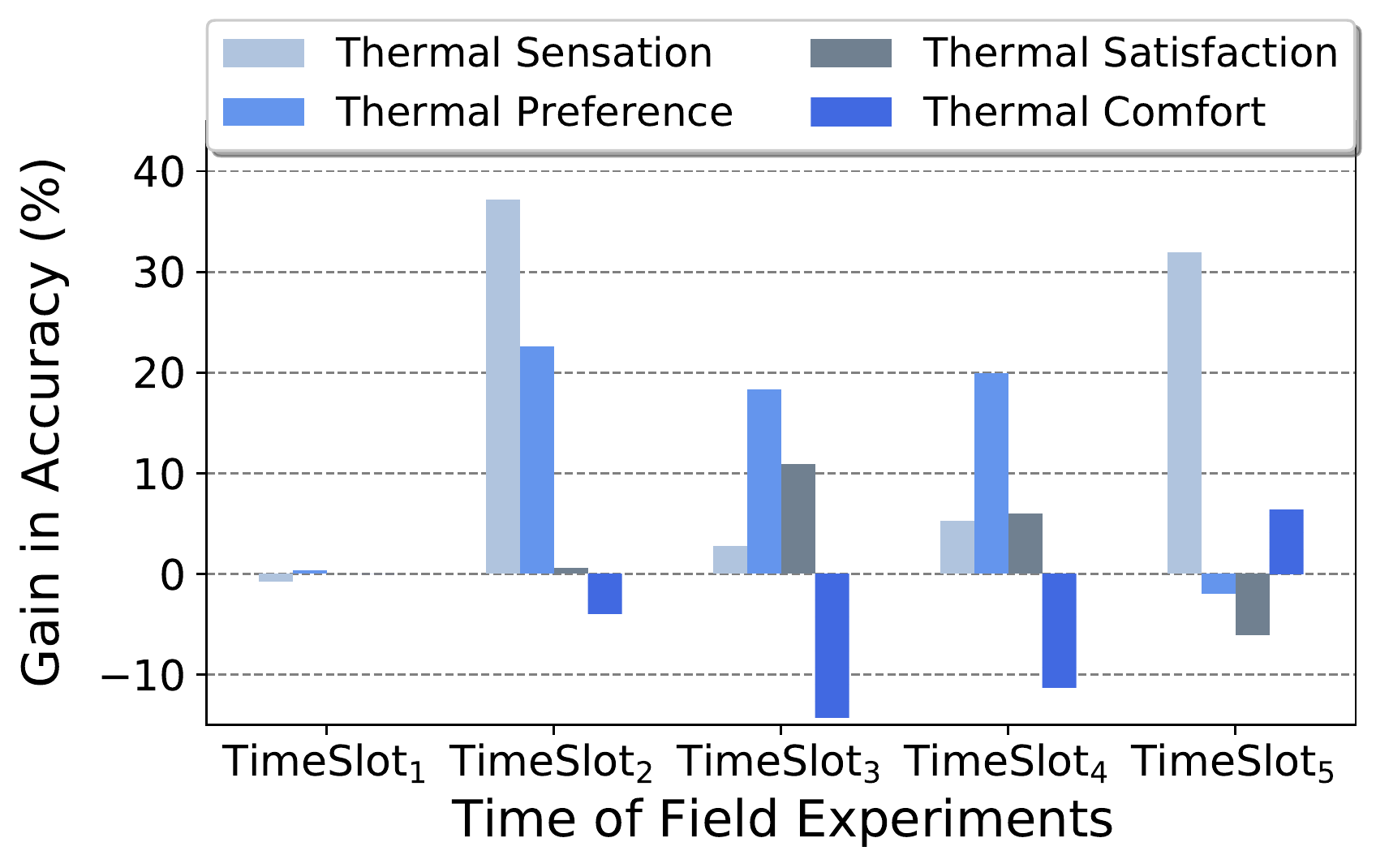}}
      \subfloat[All Time Slots for a Location] { \includegraphics[width=.33\linewidth]{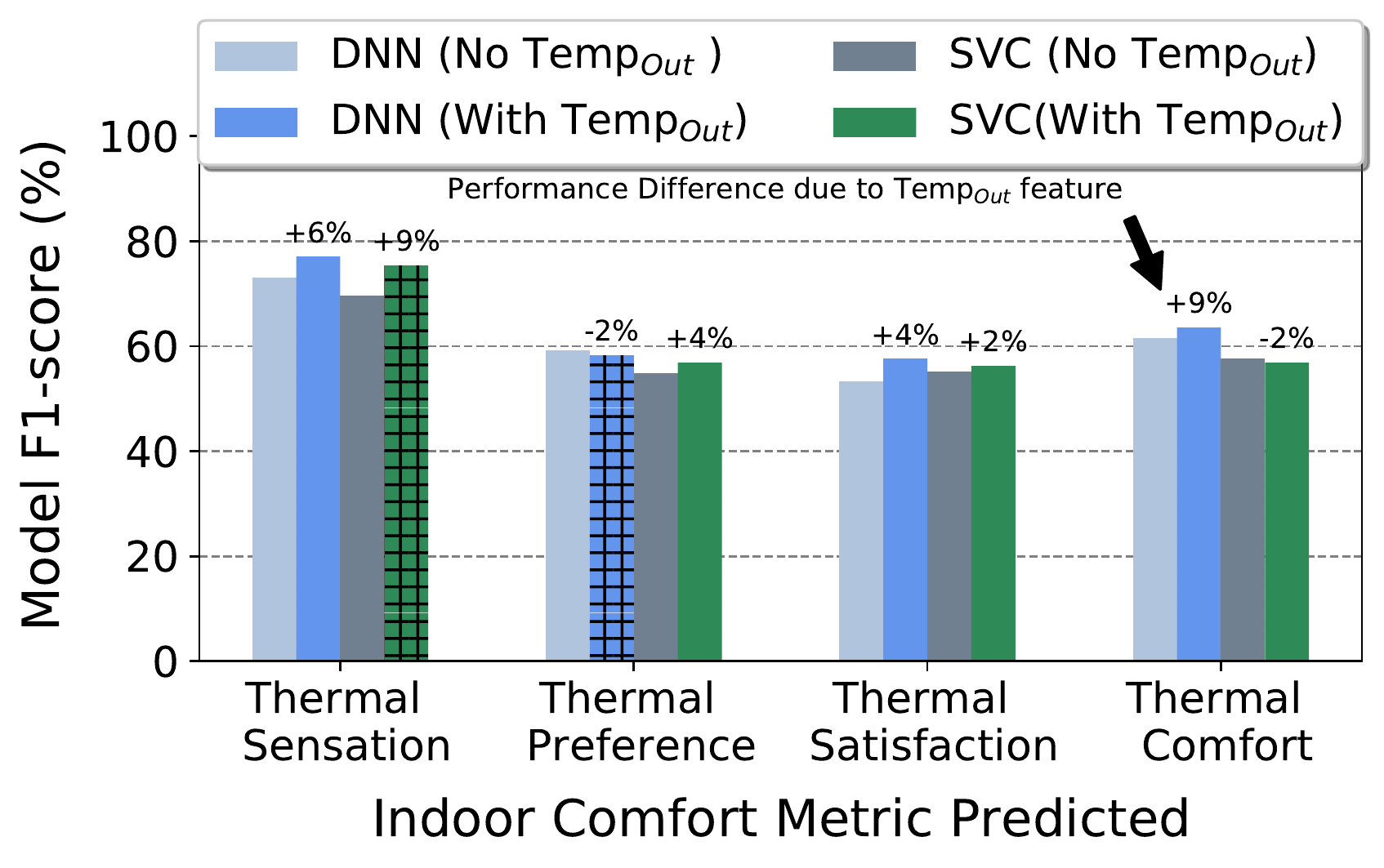}}\hspace*{0.2cm}
\end{tabular}
  \caption{Daytime variation in Outdoor temperature and TC Prediction Performance}  
    \label{fig:outdoor}
    \vspace*{-0.4cm}
\end{figure*}

\par\textbf{{Evaluation Methodology:}}
For train-test validation, \textit{stratified sampling} was employed with 5 splits, to conserve the class distribution. Further, multiple train-test ratios were considered in experiments. Performance of all DNN and SVM models is evaluated on four parameters viz., Accuracy, F1-score, Precision, and Recall. The results are averaged over the stratified splits. Although all 4 parameters are important, due to paucity of space, results with respect to maximum average Accuracy are presented. Likewise, when similar trends are observed for both SVM and DNN models, only one is presented. Classification performance is evaluated through confusion matrices, for the best performing model configuration. 


\section{Results and Analysis}
The results of the experiments for the three objectives outlined above are presented next.
\par\textbf{Impact of Light Exposure on Model Performance:}
The CIE Sky illuminance models~\cite{ISHRAE}, shown in Figure~\ref{fig:circadian}(a), show the hourly diffused solar radiation for the city of Dehradun, for 4 days (out of 28 days) during which experiments were conducted. Typically, after sun-rise, the light exposure increases as the day progresses, peaks around noon, and then reduces gradually. However, due to factors such as cloud cover intensity and opacity, fog, rain, etc., light exposure fluctuates.  It can be observed that the intensity of solar radiation and illuminance distribution  of sky fluctuates during a day and does not follow a steady rise and decline pattern. This high temporal variability in light exposure clearly impacts TC prediction model performance. 

Considering all the time slots, the model Accuracy results in Figure~\ref{fig:circadian}(b), show an interesting trend. For all four TC metrics, the model performance generally improves as time of day or light exposure increases. There are certainly some expected fluctuations based on the discussion above. For example, TimeSlot\textsubscript{3}, has the best performance. Nevertheless, there is a \textit{definitive impact of time of day and circadian rhythms on the prediction performance of subjective TC perceptions}. 
\par\textbf{Temporal Specificity of TC Prediction Models:}
The multi-class classification performance of TCV (comfort level) shown in Figure~\ref{fig:classification}, is the best indicator of time specificity of TC prediction performance, as it demonstrates a high temporal variation. 
It's evident that mislabeling of class values reduces drastically from TimeSlot\textsubscript{2} to TimeSlot\textsubscript{5}, i.e., as the time of day increases. Consequently, the model performance improves significantly, with respect to Accuracy and F1-scores, by as much as 100\% and 83\%, respectively. Further, the temporal specificity can be observed in both majority and minority classes, which implies that \textit{taking ``time of day" into consideration may help address the challenge of class imbalance.}
Thus, classification performance seems to be highly temporal, depending upon the photic input and outdoor thermal environment that students experience. 

Although, the model for All Time Slots (Figure~\ref{fig:classification}(a)) does not perform very well for TCV, it demonstrates average performance for other TC Metrics as shown in Figure~\ref{fig:outdoor}(a). Thus, train-once-test-any-time strategy may be sufficient for some TC metrics, especially TSV, but metrics such as TCV may need a time-specific models.

\par\textbf{Impact of Outdoor Temperature:} Adding the new outdoor temperature feature to the DNN and SVM models trained on the baseline primary student dataset~\cite{bettydeepcomfort, bettySmartsys} significantly improves the TC prediction performance for all time slots combined. This is evident from Figure~\ref{fig:outdoor}(a) and is true for all four TC metrics considered. 

However, for individual time slot scenario, i.e., trained on all data and tested on individual time slot data, the results vary in each time slot. On the whole,Figure~\ref{fig:outdoor}(b) shows that inclusion of outdoor temperature improves the prediction performance for all metrics except TCV. The most prominent gains are in TSV and TPV. 
Surprisingly, TCV model performance drops when outdoor temperature is included. We have earlier learned that primary students find it extremely difficult to judge their ``comfort levels", leading to a high number of `\textit{illogical votes}~\cite{bettydeepcomfort, bettySmartsys, bettySust}. Consequently, TCV prediction models for children perform far worse than those for adults~\cite{bettySmartsys}. Here, a probable explanation is that among all time slots, TimeSlot\textsubscript{3} (Figure~\ref{fig:scales}(c)) has the lowest fraction of students who are slightly comfortable (6\%) and the highest percentage of very comfortable students (34\%). Looking at the corresponding TPV values (Figure~\ref{fig:scales}(b)), close to 45\% students want the classroom to be ``warmer" or ``much warmer." Thus, adding outdoor temperature and the time of day features, seems to be confusing the classifier.
\par\textbf{Spatio-temporal Variation in TC Prediction:} School\textsubscript{1} is considered for location specific temporal variability analysis, as it is the best site in our study with respect to number of unique students (103), duration of experiments/classroom (5 days on avg.), and temporal span (5 time slots), contributing to 25\% of data samples.
With respect to the time of day and light exposure, spatial context seems to have some impact, which is visible in Figure~\ref{fig:circadian}(c). Ignoring TimeSlot\textsubscript{6}, which has only 40 samples, the prediction model Accuracy gradually increases as the time approaches noon. However, this is not true for TSV (sensation), where TimeSlot\textsubscript{1} shows highest prediction Accuracy. This is an extremely important finding, as the natural illumination the classrooms of School\textsubscript{1} was excellent in the morning slots owing to the orientation with respect to sun and an unobstructed layout. Further, the average outdoor temperature for the days of the experiments was higher (16.04°C) than all other schools (13.09°C, 14.4°C, 13.6°C, and 14.9°C). 

Finally, let us consider the localized temporal variation in outdoor temperature, and its impact on TC prediction. The results are presented in Figure~\ref{fig:outdoor}(c). The trends are similar for both DNN and SVM, indicating no overfiting in the former. Neural network models perform better overall. For most TC metrics, there is a gain in F1-scores with the inclusion of the running mean outdoor temperature as a feature. However, the increase is subdued as compared to the macro scenario in Figure~\ref{fig:outdoor}(a).
Thus, when considering TC prediction models at the micro level, i.e., for specific buildings, the spatial variability must also be taken into account along with temporal variability. 

\par The important inferences from the above discussion are presented in the next section along with the future direction of this work.




\section{Conclusions and Future Work}~\label{sec:conclusions}
This work tried to answer three important research problems concerning the impact of temporal variability on thermal comfort prediction.
There were several new findings, the most insightful among which are:
\par1. With increase in time of day, thermal comfort prediction performance generally improves.
\par2. Temporal variation in outdoor temperature has a significant positive impact on TC prediction, overall. However, at specific time of the day, the temporal context determines the magnitude of this impact.
\par3. Train-once-test-anytime models are possible at the macro level (e.g., a group of schools or residential buildings) for some thermal comfort metrics (e.g., thermal sensation).
\par4. For precise micro-level (e.g., specific building or indoor space) prediction capabilities, features governing spatio-temporal variability must be included in the model.
\par5. Limited cognition and comprehension abilities of children introduce complexities in TCV (comfort) and TPV (preference) prediction. 

We intend to extend this work by predicting the comfort temperature with high accuracy in different temporal contexts.
Further, we aim to test the generalization ability of the ML prediction models, across \textit{hours}, \textit{days}, and \textit{seasons}. 
For contexts with limited samples such as TimeSlot\textsubscript{6}, solutions such as transfer learning will be explored. 



\section{Acknowledgment}
The research was funded by the Sasakawa Scientific Research Grant of the Japan Science Society and JSPS KAKENHI Grant Number JP 22H01652. The authors are grateful to the administrators of 
the five schools 
and Mrs. Pushpa Manas, (Retd.) Director of School Education, Uttarakhand, India, for facilitating this study.

\bibliography{ref, rel, MLTCR1, MLTCR2}

\begin{thebibliography}{10}

\bibitem{[2]}
B.~P.M, {\em The Indoor Environment Handbook: How to make buildings healthy and
  comfortable}.
\newblock London, UK: Earthscan.

\bibitem{Fanger}
P.~O. Fanger {\em et~al.}, ``Thermal comfort analysis and applications in
  environmental engineering.,'' {\em Thermal comfort. Analysis and applications
  in environmental engineering.}, 1970.

\bibitem{Adaptive}
R.~De~Dear and G.~S. Brager, ``Developing an adaptive model of thermal comfort
  and preference,'' 1998.

\bibitem{ML_TC_REVIEW_1}
L.~{Arakawa Martins}, V.~Soebarto, and T.~Williamson, ``A systematic review of
  personal thermal comfort models,'' {\em Building and Environment}, vol.~207,
  p.~108502, 2022.

\bibitem{ML_TC_REVIEW_2}
Z.~{Qavidel Fard}, Z.~S. Zomorodian, and S.~S. Korsavi, ``Application of
  machine learning in thermal comfort studies: A review of methods, performance
  and challenges,'' {\em Energy and Buildings}, vol.~256, p.~111771, 2022.

\bibitem{ventilation}
{Noack, Rick and Hassan, Jennifer}, 2019.

\bibitem{bettySmartsys}
B.~Lala, S.~M. Kala, A.~Rastogi, K.~Dahiya, H.~Yamaguchi, and A.~Hagishima,
  ``Building matters: Spatial variability in machine learning based thermal
  comfort prediction in winters,'' in {\em 2022 IEEE International Conference
  on Smart Computing (SMARTCOMP)}, pp.~342--348, 2022.

\bibitem{ANSIASHRAE2020}
ANSI/ASHRAE, ``Ansi/ashrae standard 55-2020: thermal environmental conditions
  for human occupancy,'' 2020.

\bibitem{timeold1}
F.~H. Rohles~Jr, ``The effect of time of day and time of year on thermal
  comfort,'' in {\em Proceedings of the Human Factors Society Annual Meeting},
  vol.~23, pp.~129--132, SAGE Publications Sage CA: Los Angeles, CA, 1979.

\bibitem{timeold2}
P.~Fanger, J.~H{\o}jbjerre, and J.~Thomsen, ``Thermal comfort conditions in the
  morning and in the evening,'' {\em International journal of biometeorology},
  vol.~18, no.~1, pp.~16--22, 1974.

\bibitem{timenew1_18}
N.~Kakitsuba, Q.~Chen, and Y.~Komatsu, ``Diurnal change in psychological and
  physiological responses to consistent relative humidity,'' {\em Journal of
  Thermal Biology}, vol.~88, p.~102490, 2020.

\bibitem{VelleitimeAnalysis}
M.~Vellei, W.~O’brien, S.~Martinez, and J.~Le~Dr{\'e}au, ``Some evidence of a
  time-varying thermal perception,'' {\em Indoor and Built Environment},
  p.~1420326X211034563, 2021.

\bibitem{VelleitimeSurvey}
M.~Vellei, G.~Chinazzo, K.-M. Zitting, and J.~Hubbard, ``Human thermal
  perception and time of day: A review,'' {\em Temperature}, vol.~8, no.~4,
  pp.~320--341, 2021.

\bibitem{TimeTemp}
Y.~Sun, X.~Luo, and H.~Ming, ``Analyzing the time-varying thermal perception of
  students in classrooms and its influencing factors from a case study in
  xi’an, china,'' {\em Buildings}, vol.~12, no.~1, p.~75, 2022.

\bibitem{timehealthbenefits_21}
Y.~M. Ivanova, H.~Pallubinsky, R.~Kramer, and W.~van Marken~Lichtenbelt, ``The
  influence of a moderate temperature drift on thermal physiology and
  perception,'' {\em Physiology \& Behavior}, vol.~229, p.~113257, 2021.

\bibitem{timeenergybenefits_20}
A.~Mishra, M.~Loomans, and J.~L. Hensen, ``Thermal comfort of heterogeneous and
  dynamic indoor conditions—an overview,'' {\em Building and Environment},
  vol.~109, pp.~82--100, 2016.

\bibitem{[59]}
G.~Havenith, ``Metabolic rate and clothing insulation data of children and
  adolescents during various school activities,'' {\em Ergonomics}, vol.~50,
  p.~1689 – 1701.

\bibitem{bettydeepcomfort}
B.~Lala, H.~Rizk, S.~M. Kala, and A.~Hagishima, ``Multi-task learning for
  concurrent prediction of thermal comfort, sensation and preference in
  winters,'' {\em Buildings}, vol.~12, no.~6, p.~750, 2022.

\bibitem{ISHRAE}
``The indian society of heating, refrigerating and air conditioning engineers
  (ishrae) illuminance data,'' 2019.
\newblock Data provided by \url{https://ishrae.in/}, downloaded from
  \url{https://energyplus.net/weather-location/asia_wmo_region_2/IND/IND_Dehradun.421110_ISHRAE},
  and rendered at \url{https://drajmarsh.bitbucket.io/cie-sky.html}
  \url{http://www.csie.ntu.edu.tw/~cjlin/libsvm}.

\bibitem{timeold3}
P.~Fanger, O.~{\"O}stberg, N.~Breum, E.~Jerking, {\em et~al.}, ``Thermal
  comfort conditions during day and night,'' {\em European Journal of Applied
  Physiology and Occupational Physiology}, vol.~33, no.~4, pp.~255--263, 1974.

\bibitem{timeold4_yes}
F.~Grivel and V.~Candas, ``Ambient temperatures preferred by young european
  males and females at rest,'' {\em Ergonomics}, vol.~34, no.~3, pp.~365--378,
  1991.

\bibitem{basal_11}
K.-M. Zitting, N.~Vujovic, R.~K. Yuan, C.~M. Isherwood, J.~E. Medina, W.~Wang,
  O.~M. Buxton, J.~S. Williams, C.~A. Czeisler, and J.~F. Duffy, ``Human
  resting energy expenditure varies with circadian phase,'' {\em Current
  Biology}, vol.~28, no.~22, pp.~3685--3690, 2018.

\bibitem{db1}
R.~J. De~Dear, ``A global database of thermal comfort field experiments,'' {\em
  ASHRAE transactions}, vol.~104, p.~1141, 1998.

\bibitem{daylightTC}
G.~Chinazzo, J.~Wienold, and M.~Andersen, ``Daylight affects human thermal
  perception,'' {\em Scientific reports}, vol.~9, no.~1, pp.~1--15, 2019.

\bibitem{ko2020impact}
W.~H. Ko, S.~Schiavon, H.~Zhang, L.~T. Graham, G.~Brager, I.~Mauss, and Y.-W.
  Lin, ``The impact of a view from a window on thermal comfort, emotion, and
  cognitive performance,'' {\em Building and Environment}, vol.~175, p.~106779,
  2020.

\bibitem{lightTC_23}
M.~te~Kulve, L.~Schellen, L.~Schlangen, and W.~van Marken~Lichtenbelt, ``The
  influence of light on thermal responses,'' {\em Acta Physiologica}, vol.~216,
  no.~2, pp.~163--185, 2016.

\bibitem{bettyIndia}
B.~Lala, ``Analysis of thermal comfort study in india,'' in {\em International
  Conference on Civil, Architecture, Environment and Waste Management
  (CAEWM-17)}, 2017.

\bibitem{bettySust}
B.~Lala, S.~Murtyas, and A.~Hagishima, ``Indoor thermal comfort and adaptive
  thermal behaviors of students in primary schools located in the humid
  subtropical climate of india,'' {\em Sustainability}, vol.~14, no.~12,
  p.~7072, 2022.

\bibitem{db2}
``Development of the ashrae global thermal comfort database ii,'' {\em
  Environ}, vol.~142, p.~502–512.

\bibitem{libsvm}
C.-C. Chang and C.-J. Lin, ``{LIBSVM}: A library for support vector machines,''
  {\em ACM Transactions on Intelligent Systems and Technology}, vol.~2,
  pp.~27:1--27:27, 2011.
\newblock Software available at \url{http://www.csie.ntu.edu.tw/~cjlin/libsvm}.

\bibitem{Adam}
D.~P. Kingma and J.~Ba, ``Adam: A method for stochastic optimization,'' {\em
  ICLR}, 2015.

\end{thebibliography}
\newpage
\begin{IEEEbiography}[{\includegraphics[width=1in,height=1.25in,clip,keepaspectratio]{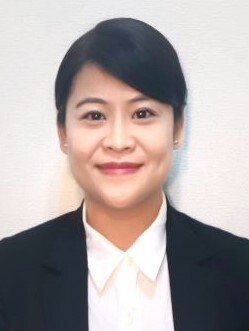}}]{Dr. Betty Lala} is a doctoral researcher at the Interdisciplinary Graduate School of Engineering Sciences (IGSES), Kyushu University, Japan. Prior to this, she was an Assistant Professor in the Faculty of Architecture, Manipal University, India. She was also a professional Architect for 3 years specializing in villas and luxury high-rise buildings. She received her masters degree in Architecture from IIT Roorkee, India. She was a recipient of the DAAD scholarship, and carried out her master's research at TU Berlin, Germany. She also received the prestigious Sasakawa scientific research grant in 2020-21. Her interests lie in the field of thermal comfort, research survey methodology, data analysis, urban design, and UX design.
\end{IEEEbiography}

\begin{IEEEbiography}[{\includegraphics[width=1in,height=1.25in,clip,keepaspectratio]{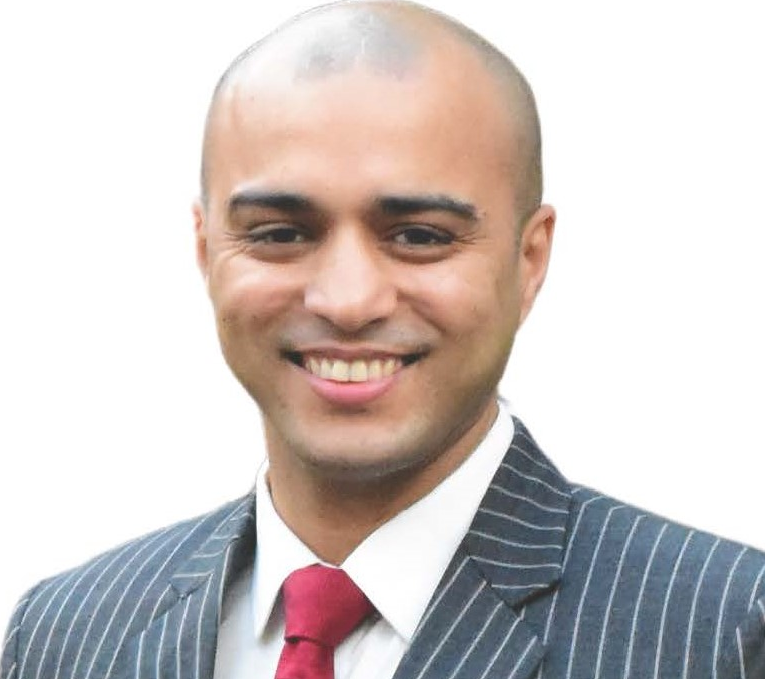}}]{S{rikant}  M{anas} K{ala}} is a doctoral researcher at the Mobile Computing Lab, Osaka University, Japan. He is also a VC Global Analyst with Plug and Play Tech Center. He received his M.Tech degree in Computer Science and Engineering from IIT Hyderabad, India. He has been awarded the Employee Excellence Award by Infosys and IIT Hyderabad Research Excellence Award in 2016 and 2017. He led his startup team to the semifinals of Ericcson Innovation Awards 2020 and the Impact Summit of Hult Prize 2021.  His interests lie in the domain of Extended Reality, applied AI/ML, Venture Capital investment analysis, indoor thermal comfort, and Unlicensed and 5G Networks,
\end{IEEEbiography}
\begin{IEEEbiography}[{\includegraphics[width=1in,height=1.25in,clip,keepaspectratio]{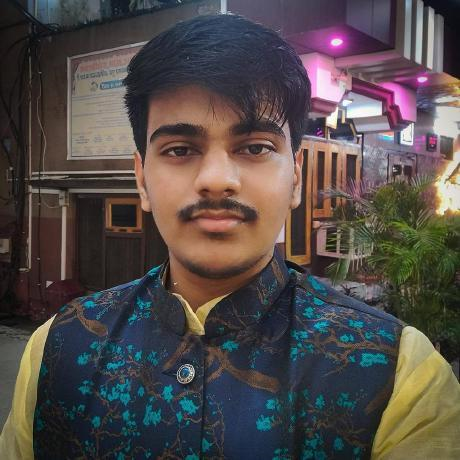}}] {A{nmol} R{astogi}} is a final year undergraduate student in the department of Artificial Intelligence at IIT Hyderabad, India. His research interests lie in the field of machine learning and deep learning.
\end{IEEEbiography}

\begin{IEEEbiography}[{\includegraphics[width=1in,height=1.25in,clip,keepaspectratio]{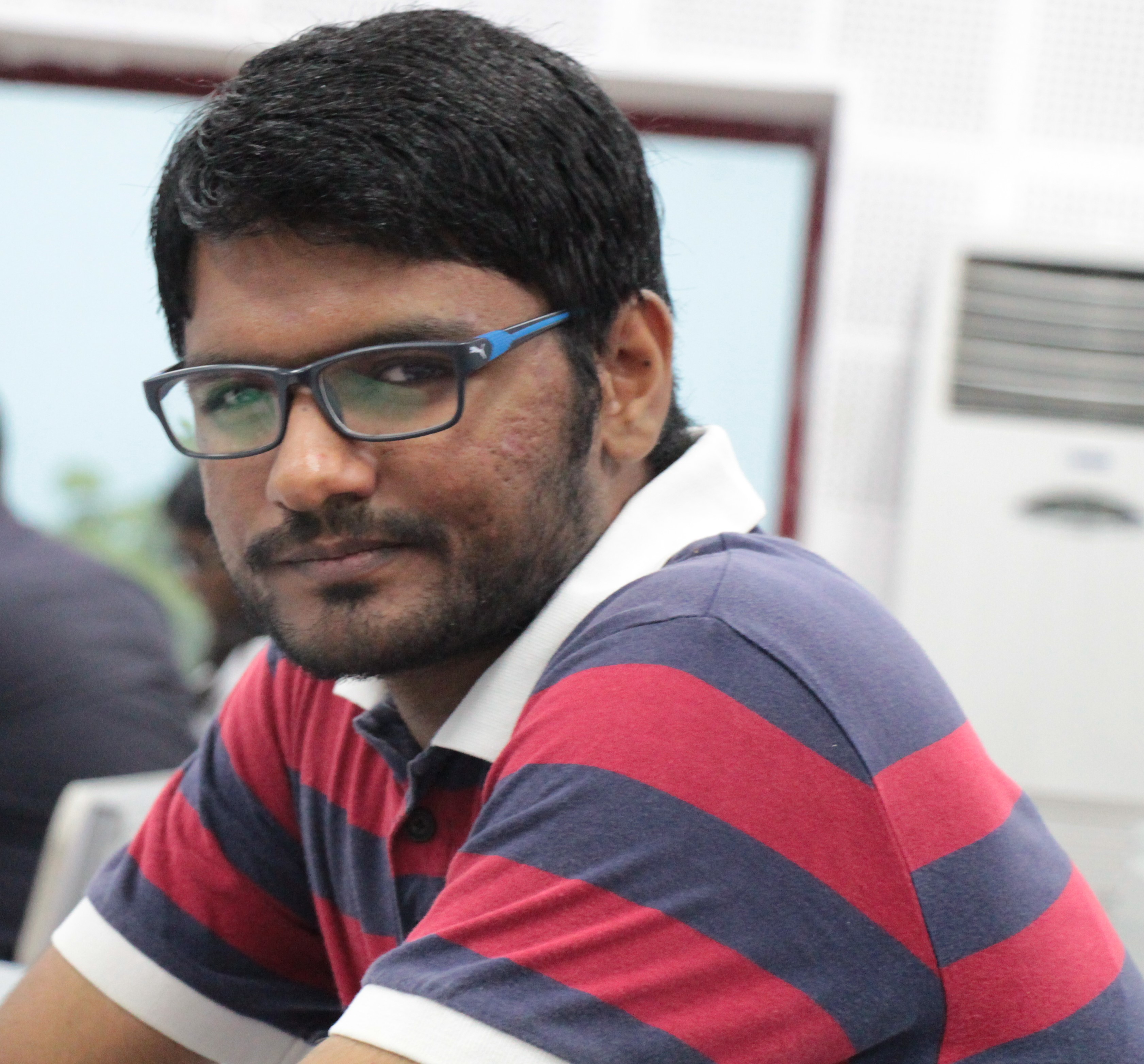}}] {K{unal} D{ahiya}} is currently a Research Scholar at IIT Delhi and Research Intern at Microsoft Research India, where he works on deep extreme multi-label learning. His work has not only led to publications in leading conferences like ICML, CVPR, and WSDM but has found applications in various real-world applications, including query recommendations and ads benefiting millions of users and small businesses. He received his B.Tech and M.Tech degrees from IIT Hyderabad, where he worked on large-scale visual computing applications. His interests lie in extreme multi-label learning, Siamese networks, representation learning, imbalanced classification, 5G and LAA Network Operator Data Analysis, and thermal comfort prediction.
\end{IEEEbiography}

\begin{IEEEbiography}[{\includegraphics[width=1in,height=1.25in,clip,keepaspectratio]{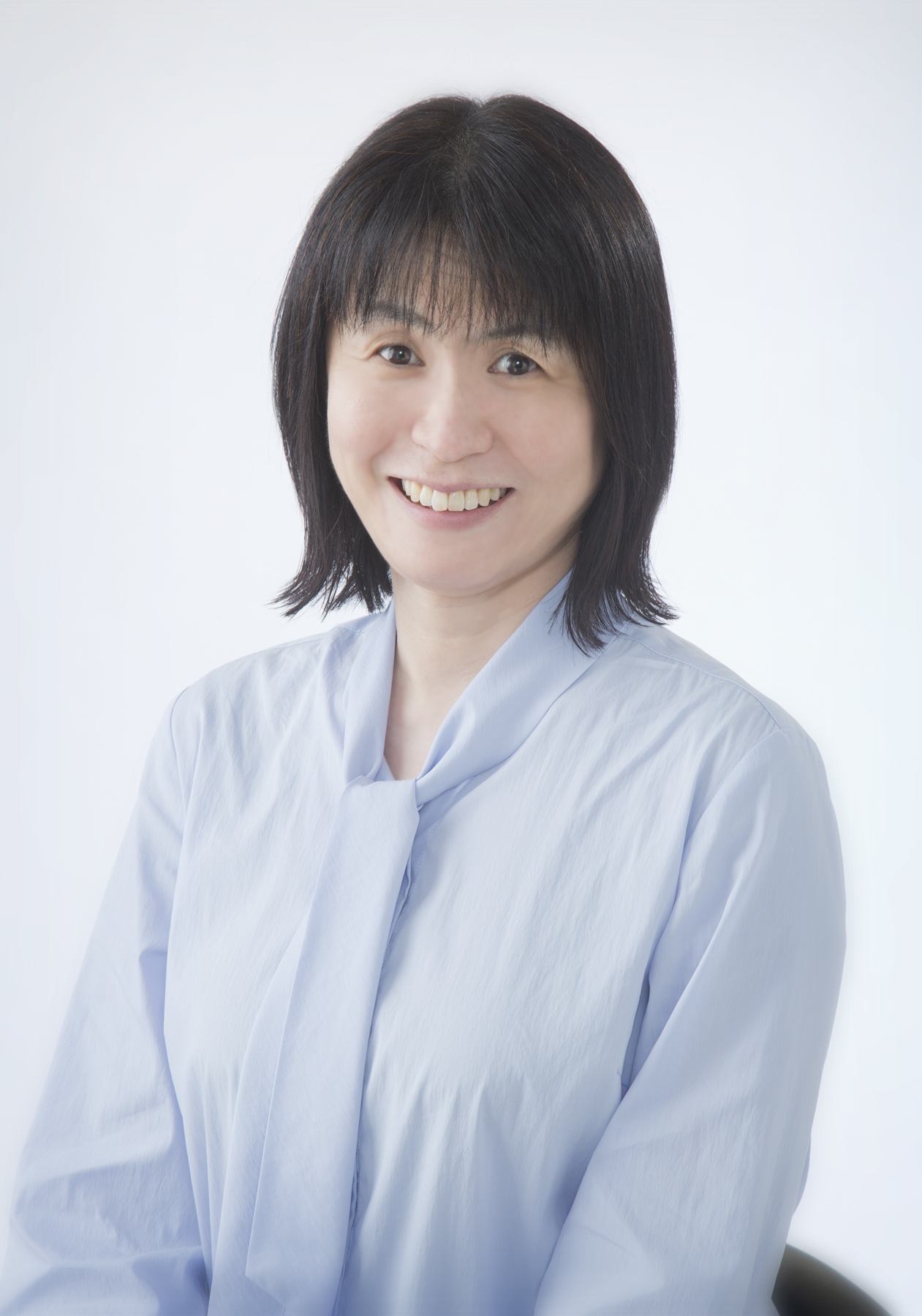}}]{D{r}. A{ya} H{agishima}} received the Dr. Eng degree from the Interdisciplinary Graduate School of Engineering Sciences (IGSES), Kyushu University, Fukuoka, Japan, in 2005. She is currently a Professor in the Faculty of Engineering Sciences, Kyushu University. She has worked mainly in the research areas of urban climatology and building environment.  Her current research interests include zero-energy buildings/houses, energy-related occupants behaviours, and sustainable built environment.
\end{IEEEbiography}
\end{document}